\documentclass[sigconf]{acmart}

\usepackage{xcolor}
\definecolor{mutedblue}{HTML}{5F88D8}
\usepackage{amsmath, amsfonts}
\usepackage{booktabs}
\usepackage{microtype}
\usepackage{xcolor}
\usepackage[table]{xcolor}
\definecolor{cyan}{rgb}{0.0,0.8,0.8}
\usepackage{url}
\usepackage{nicefrac}
\usepackage{graphicx}
\usepackage{graphics}
\usepackage{algorithmic}
\usepackage{algorithm}
\usepackage{multirow}

\usepackage{pifont}
\usepackage{caption}
\usepackage{subcaption}
\usepackage{makecell}

\usepackage{xcolor, colortbl}
\definecolor{scarlet}{rgb}{1, 0.85, 0.85}
\usepackage[toc,page]{appendix}
\usepackage{caption}
\usepackage{subcaption}
\usepackage{amsthm}

\newcommand{\appsection}[1]{
  \refstepcounter{section}
  \section*{Appendix \Alph{section}: #1}
  \addcontentsline{toc}{section}{Appendix \Alph{section}: #1}
}
\usepackage{newfloat}
\usepackage{listings}
\DeclareCaptionStyle{ruled}{labelfont=normalfont,labelsep=colon,strut=off}
\floatstyle{ruled}
\newfloat{listing}{tb}{lst}{}
\floatname{listing}{Listing}
\definecolor{revblue}{RGB}{0,0,255}
\newcommand{\rev}[1]{#1} 

\AtBeginDocument{%
  }

\copyrightyear{2026}
\acmYear{2026}
\setcopyright{cc}
\setcctype{by-nc-nd}
\acmConference[CIKM '26]{Proceedings of the 35th ACM International Conference on Information and Knowledge Management}{November 07--11, 2026}{Rome, Italy}
\acmBooktitle{Proceedings of the 35th ACM International Conference on Information and Knowledge Management (CIKM '26), November 07--11, 2026, Rome, Italy}
\acmDOI{10.1145/3799682.3840726}
\acmISBN{979-8-4007-2539-5/2026/11}
\begin{document}

\title{Simple Actors and Deep Critics for Scalable Reinforcement Learning}

\author{Guhyeon Kang}
\orcid{0009-0000-6005-5909}
\affiliation{%
  \institution{Sungkyunkwan University}
  \country{Republic of Korea}
}
\email{guhyeon.kang@skku.edu}

\author{Jaehwi Lee}
\orcid{0009-0001-8014-6493}
\affiliation{%
  \institution{Soongsil University}
  \country{Republic of Korea}
}
\email{jaehwilee@soongsil.ac.kr}

\author{Minhae Kwon}
\authornote{Corresponding author. G. Kang and M. Kwon are with the Department of Electrical and Computer Engineering, Sungkyunkwan University, Republic of Korea. J. Lee is with Soongsil University, Republic of Korea.}
\affiliation{%
  \institution{Sungkyunkwan University}
  \country{Republic of Korea}
}
\email{minhae.kwon@skku.edu}

\begin{abstract}
Recent progress in offline reinforcement learning (RL) has been driven by
expressive generative actors such as diffusion and flow-matching
policies, which capture multimodal behavior in offline datasets.
However, these actors require multiple denoising or integration
steps per action and thus incur substantial overhead at every
decision in deployment. In this work, we revisit where capacity
should be invested in an offline actor--critic method. Since the critic is used only during training
and is discarded at deployment while the actor runs at every decision
step, allocating capacity to the critic rather than the actor is more
favorable for inference-time efficiency. However, scaling MLP critics in offline RL is known to introduce several distinct instabilities that have, in practice, kept critics shallow. We identify three
distinct failure modes that arise when critics are deepened in offline
RL---optimization, bootstrap-noise amplification, and value-range
drift---and address each with a corresponding ingredient: a residual
MLP backbone, $n$-step bootstrap targets, and a categorical
cross-entropy loss. Combining these ingredients with a lightweight
deterministic actor, we propose \textbf{LAC} (\textbf{L}ight \textbf{A}ctor, deep \textbf{C}ritic). On OGBench, LAC matches the strongest diffusion- and flow-matching baselines while achieving up to $4\times$ lower inference latency, comparable to one-step distilled policies without distillation. Its critic recipe also transfers across actor parametrizations.\

\smallskip\noindent\textbf{Project page:} \url{https://9hyeon1225.github.io/LAC/}
\end{abstract}



\maketitle


\section{Introduction}\label{sec:introduction}
Generative actors---diffusion-based~\cite{wang2023diffusionql,chi2023diffusionpolicy}
and flow-matching policies~\cite{park2025fql,li2025qchunking,hansen2023idql}---are
reshaping offline RL and increasingly serve as the
backbone of agentic systems for robotics and physical artificial intelligence~\cite{black2025pi0,kim2024openvla,davis2024airbnb,agrawal2023rlqr,arya2025relink, lee2024ad4rl, eom2026price, lee2025episodicdriving,lee2022adasrl}, as well as for information-driven applications such as recommendation~\cite{chen2025dac4rec,chen2025mdt4rec}.
Their strength is representational: they capture multimodal action
distributions that Gaussian policies cannot, which translates into stronger
performance on diverse offline datasets. However, this gain comes with a
structural cost that is rarely interrogated. Each action requires multiple denoising or integration steps 
through a high-capacity network, so every decision at deployment 
incurs the full cost of the actor architecture, which accumulates 
in resource-constrained deployments such as robotics and embedded 
control. The community has
implicitly accepted this as the price of expressivity, but it raises a
question that has not been put sharply enough.
\begin{center}
\textit{Is the architectural budget in offline actor--critic methods allocated
to the right network?}
\end{center}

\begin{figure}[t]
    \centering
    \includegraphics[width=\columnwidth]{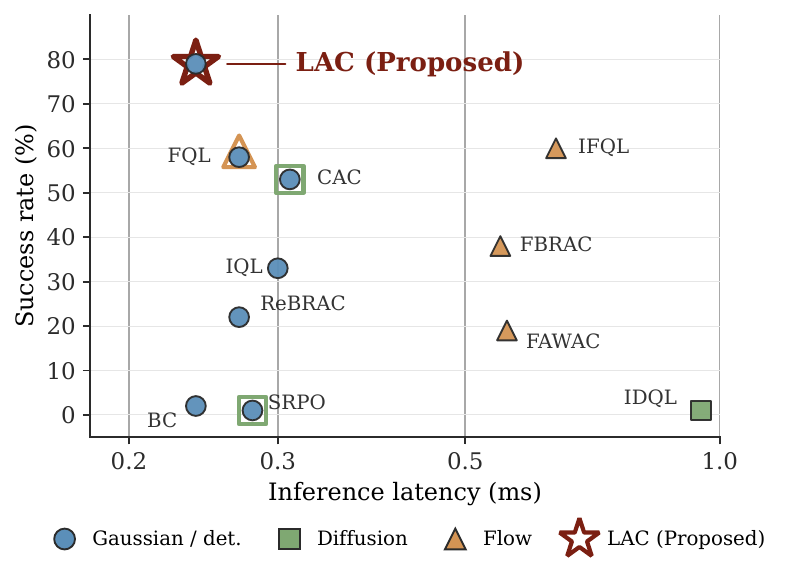}
    \vspace{-1.5em}
    \caption{Performance vs.\ inference latency on
    \texttt{humanoidmaze-medium}. Iterative generative actors
    (\textsc{IDQL}, \textsc{IFQL}, \textsc{FBRAC}, \textsc{FAWAC}) require multiple
    denoising or flow-integration steps per action.\protect\footnotemark  
    \textbf{LAC} (star) acts in a single forward pass, reaching higher success rate at lower latency.}
    \label{fig:intro}
\vspace{-1.0em}
\end{figure}
 
\footnotetext[1]{The overlapping markers for \textsc{SRPO}, 
\textsc{CAC}, and \textsc{FQL} indicate that they leverage diffusion- 
or flow-based generative models during training to derive 
single-step Gaussian or deterministic actors used at inference.}

We revisit this assumption. The actor and the critic occupy fundamentally asymmetric roles across the training--deployment boundary: the critic exists only to shape the policy gradient during training and is \emph{discarded} once training ends, whereas the actor must run at every decision step in the deployed system. Capacity invested in a generative actor therefore pays a price that recurs at every action, while capacity invested in the critic pays it only once. Such deployment-time computational costs are particularly consequential for resource-constrained edge and IoT systems, where lightweight real-time inference is a key design consideration~\cite{kim2025psfl}. This suggests a complementary direction to the current generative actor trend: rather than scaling the network that must run at inference, scale the network that is free at inference. Concretely, pair a \emph{lightweight deterministic} actor with a \emph{deep and well-trained} critic, allowing the critic to provide a stronger learning signal while keeping the deployed policy lightweight. A natural concern is whether a simple deterministic actor can remain competitive with expressive generative actors on datasets with multimodal behavior. We show that, when guided by a sufficiently strong deep critic, a lightweight deterministic actor can remain competitive across standard offline RL benchmarks.

This asymmetric allocation is not free to realize. Simply increasing the
depth of an MLP critic is known to destabilise value-based RL, which is part
of why offline RL has traditionally kept critics
shallow~\cite{sinha2020d2rl,bjorck2021deeperrl,peng2024deadly}.
We attribute this to three distinct failure modes. The first is an
\emph{optimization} problem: gradients fail to propagate through many
layers. The second is a \emph{bootstrap-noise} problem: a deep critic with
high representational capacity also has high capacity to fit noisy
predictions, and the $1$-step bootstrap update folds those predictions back
into the target at every iteration, causing errors to compound. The third
is a \emph{value-range drift} problem: even when bootstrap noise is
suppressed, a deep critic trained with unbounded Mean Squared Error (MSE) regression can slowly
drift its outputs outside any reasonable scale, at which point a single
anomalous sample is enough to push training into divergence---often after
the run has already appeared to converge. Standard ingredients address only
one of these axes at a time. We therefore combine three ingredients along
complementary axes:
(i) a residual MLP backbone with layer normalization, which restores
gradient flow~\cite{nauman2024bro,doro2023bbf,sinha2020d2rl};
(ii) $n$-step bootstrap targets~\cite{sutton2018reinforcement,kapturowski2019r2d2,hessel2018rainbow},
which incorporate $n$ observed rewards and change the bootstrap coefficient
from $\gamma$ to $\gamma^n$, thereby reducing reliance on the critic's own
prediction, and
(iii) a categorical cross-entropy critic loss over a bounded
support~\cite{bellemare2017c51,farebrother2024stopregression,imani2018hlgauss}, which
restricts target values to a fixed range and converts regression into a
classification problem over that range. Together, the three ingredients
let the critic scale stably across two orders of magnitude in depth,
providing a learning signal strong enough that a lightweight deterministic
actor suffices. Critically, we find that $n$-step is not a generic
performance booster but an enabling condition for depth, and that the
categorical loss is not interchangeable with MSE regression even when
$n$-step is in place: removing either ingredient causes a distinct,
depth-specific failure mode (Section~\ref{sec:exp_capacity}).
Figure~\ref{fig:intro} shows the result of putting all three together:
matched against expressive flow-matching and diffusion baselines, our
design recovers comparable performance while reducing per-action inference
cost by up to $4\times$\rev{ over multi-step generative actors}.
 
We instantiate these ideas in \textbf{LAC}, a simple offline
actor--critic recipe that integrates: (i) a deep residual MLP critic with
layer normalization, (ii) a categorical cross-entropy critic loss with
bounded support paired with $n$-step bootstrap targets, and (iii) a lightweight deterministic actor trained via the
deterministic policy gradient with a behavior-cloning regularizer, so that
producing an action at deployment requires only a single forward pass of a
small network.
 
Our main contributions are as follows.
1) We articulate an asymmetry in offline actor--critic methods that is
implicit in the current literature: the critic is unused at deployment,
which makes critic-side capacity essentially free at inference and motivates
a complementary alternative to scaling generative actors.
2) We identify three distinct failure modes of deep critics in offline RL---an optimization problem, a bootstrap-noise problem, and a value-range
drift problem---and show that addressing all three requires residual
architecture, $n$-step bootstrap targets, and a categorical critic loss in
combination; we directly evidence the bootstrap-noise mechanism via the
interaction between $n$ and critic depth, and the value-range mechanism via
the mid-training collapse pattern that emerges when MSE regression replaces
the categorical loss.
3) We pair this deep critic with a lightweight deterministic actor trained
via the deterministic policy gradient, eliminating the multi-step generative
sampling that diffusion and flow-matching actors require at inference.
4) On the OGBench offline RL benchmarks, \textbf{LAC}
matches diffusion- and flow-matching-based baselines while achieving up to
$4\times$ lower per-action inference latency \rev{than multi-step
generative actors}, with ablations \rev{showing that critic capacity can substantially compensate for limited actor expressivity}. To the
best of our knowledge, this is the first work to study asymmetric capacity
allocation as a deliberate design axis for inference-efficient offline RL.

\section{Related Work}
\label{sec:related}

\paragraph{\textbf{Generative Actors in Offline RL}}
Recent offline RL has increasingly relied on expressive generative
parametrizations of the policy to capture multimodal action
distributions in behavior data. Diffusion-based actors include
Diffusion-QL~\cite{wang2023diffusionql},
IDQL~\cite{hansen2023idql},
and SRPO~\cite{chen2024srpo}. To reduce the multi-step sampling
cost of diffusion policies, two alternative classes of generative
parametrizations have been explored. The first replaces the
diffusion model with a consistency model that directly maps noise
to actions in a single step, as in
CPQL~\cite{chen2024cpql} and CAC~\cite{ding2024cac}. The second uses flow matching to learn a deterministic velocity
field, as in FQL~\cite{park2025fql}, which further distills the flow
into a single-step policy at deployment. \rev{Shortcut
models~\cite{frans2024shortcut,espinosadice2025sorl} and
distillation-free diffusion objectives~\cite{chen2024dtql} obtain
one-step policies without a separate distillation stage.} While these approaches
reduce the sampling cost of expressive generative policies, they
still place the burden of modeling multimodal action distributions
on the actor executed at every decision step\rev{---a cost that compounds when
the policy is invoked repeatedly within hierarchical or agentic decision
pipelines~\cite{park2026multi2}}.

We pursue a complementary direction: rather than make expressive
actors cheaper, we ask whether expressive actors are necessary at
all when the critic is sufficiently strong.

\paragraph{\textbf{Critic Stabilization and Distributional Value Learning}}
The ingredients in the LAC critic have precedents in isolation.
Residual MLP backbones and layer-normalized residual stacks have
been studied for scaling online deep
RL~\cite{sinha2020d2rl,bjorck2021deeperrl,nauman2024bro}.
The $n$-step bootstrap is a standard variance-bias trade-off
device~\cite{sutton2018reinforcement} and is a core ingredient in
Rainbow~\cite{hessel2018rainbow} and
R2D2~\cite{kapturowski2019r2d2} for online RL. Distributional value learning~\cite{bellemare2023distributional}, in particular the categorical
formulation of C51~\cite{bellemare2017c51}, replaces the standard
MSE regression target with a cross-entropy loss over a bounded
support and has been shown to stabilize value learning in both
online and offline settings. Our contribution is to identify that
none of these ingredients is individually sufficient in the offline
regime, with each addressing a distinct failure mode of deep
critics, and that their combination is what allows critic depth to
be scaled stably in offline RL.

\paragraph{\textbf{Capacity Scaling in Reinforcement Learning}}
A growing line of work on capacity scaling in online RL includes
BBF~\cite{doro2023bbf} and recent studies pushing networks to
hundreds of layers~\cite{nauman2024bro}. Offline RL, by
contrast, has historically kept both actor and critic small, with
most published methods using two- or three-layer MLPs on both
sides. Where offline work has invested capacity, it has done so
primarily on the actor side via diffusion or flow-matching
parametrizations for expressivity reasons. We are not aware of
prior offline RL work that frames the allocation of capacity
between the actor and the critic as a deliberate design axis, or
that empirically demonstrates that critic-side capacity can compensate for limited actor-side expressivity under matched inference
budgets.

\section{Preliminaries}
\label{sec:prelim}

\paragraph{Markov Decision Process (MDP)}
An RL problem can be formulated using an MDP, which is defined as a
tuple $\mathcal{M} = \langle \mathcal{S}, \mathcal{A}, \rho, r, \gamma
\rangle$. Herein, $s_t \in \mathcal{S}$ is a state, $a_t \in
\mathcal{A}$ is an action, $\rho(s_{t+1} \mid s_t, a_t): \mathcal{S}
\times \mathcal{A} \to \Delta(\mathcal{S})$ is a state transition
probability, $r_t = R(s_t, a_t, s_{t+1}): \mathcal{S} \times
\mathcal{A} \times \mathcal{S} \to \mathbb{R}$ is a reward function,
and $\gamma \in [0, 1)$ is a temporal discount factor. The main
objective of RL is to learn a policy $\pi: \mathcal{S} \to
\Delta(\mathcal{A})$ that maximizes the expected discounted return as
follows.
\begin{equation}
\label{eq:rl_objective}
\mathcal{J}(\pi) = \mathbb{E}_{\pi}\bigg[ \sum_{t \ge 0} \gamma^{t} r_t \bigg]
\end{equation}
Standard deep RL algorithms estimate this objective via value 
functions parameterized as deep networks~\cite{mnih2015human}.

\paragraph{Offline RL}
An offline paradigm aims to learn a policy $\pi$ by leveraging a fixed
dataset $\mathcal{D} = \{(s_t, a_t, r_t, s_{t+1})\}_{t=1}^{N}$
collected by an arbitrary behavior policy, without further interaction
with the environment. The absence of additional exploration introduces
distributional shift between $\pi$ and the data-collecting policy,
which is the central technical difficulty of offline RL~\cite{fujimoto2019offpolicy,lee2025tdta,lee2023foresighted,bian2024dcminer,xin2023drl4ir,zhang2024roler,wang2024causal}.

\paragraph{Actor--Critic Methods}
A widely used approach to learning $\pi$ in offline RL is the
actor--critic framework~\cite{lillicrap2016continuous,fujimoto2018addressing,lee2025scenariofree,eom2024selective,lee2024episodicmarl}, which factorizes the problem into two
networks: an actor and a critic. Specifically, the critic
$Q_\theta(s, a)$ approximates the action-value function and predicts
the expected return of executing action $a$ at state $s$ and following
$\pi$ thereafter, while the actor $\pi_\phi(s)$ denotes the parameterized
policy. The critic is trained by minimizing the
temporal-difference (TD) loss as follows.
\begin{equation}
\label{eq:critic_loss_prelim}
\mathcal{L}(\theta) = \mathbb{E}_{(s_t, a_t, r_t, s_{t+1}) \sim \mathcal{D}}
\Big[\ell\big(\mathcal{T} Q_\theta(s_t, a_t),\, Q_\theta(s_t, a_t)\big)\Big]
\end{equation}
Herein, $\ell(\cdot)$ is a regression loss, typically the MSE, and $\mathcal{T}$ denotes the Bellman backup operator. The TD
target in its standard $1$-step form is defined as follows.
\begin{equation}
\label{eq:bellman_target_prelim}
\mathcal{T} Q_\theta(s_t, a_t) = r_t + \gamma\, Q_{\theta^-}\!\big(s_{t+1},\, \pi_\phi(s_{t+1})\big)
\end{equation}
Herein, $\theta^-$ is a target network parameter that stabilizes
training and is updated using the Polyak averaging method. The actor is
trained to produce actions that the critic rates highly, i.e., the
objective of the actor is to maximize the expected critic value as follows.
\begin{equation}
\label{eq:actor_obj_prelim}
\max_{\phi}\ \mathbb{E}_{s \sim \mathcal{D}}\big[ Q_\theta\big(s, \pi_\phi(s)\big)\big]
\end{equation}
The critic supplies the learning signal used to update the actor
through \eqref{eq:actor_obj_prelim}, while the trained actor
$\pi_\phi$ is what is used to act in the environment after training.

\paragraph{Behavior-Cloning Regularization}
A direct application of \eqref{eq:actor_obj_prelim} in the offline
setting is unstable because the critic $Q_\theta$ is queried on actions
$\pi_\phi(s)$ proposed by the current actor, which can drift outside
the support of $\mathcal{D}$ and incur extrapolation errors. To
mitigate this, a behavior-cloning (BC) term is widely adopted to keep
the policy close to dataset actions, yielding the following actor
loss.
\begin{equation}
\label{eq:actor_bc_prelim}
\mathcal{L}(\phi) = -\mathbb{E}_{(s, a) \sim \mathcal{D}}
\Big[ \lambda\, Q_\theta\big(s, \pi_\phi(s)\big)\ -\ \big\|\pi_\phi(s) - a\big\|^{2} \Big]
\end{equation}
Herein, $\lambda > 0$ is a coefficient that balances return
maximization and proximity to the dataset. This template, which
explicitly penalizes deviation from dataset actions, underlies a
broad family of offline RL methods, including
TD3+BC~\cite{fujimoto2021minimalist} and
ReBRAC~\cite{tarasov2023rebrac}. Related approaches such as
IQL~\cite{kostrikov2022iql} and AWAC~\cite{nair2020awac} instead
employ advantage-weighted regression, which arises as the
closed-form solution to a Kullback–Leibler-constrained variant of the
return-maximization objective and can be viewed as an alternative
realization of the same return-maximization-with-behavior-anchoring
principle. We adopt the form in \eqref{eq:actor_bc_prelim} for our
actor in Section~\ref{sec:method}.

\paragraph{What is Used After Training}
A property of the actor--critic framework that we exploit in this work
is that the critic $Q_\theta$ is essential during training to shape
the actor via
\eqref{eq:actor_obj_prelim} and \eqref{eq:actor_bc_prelim}, but only the
actor $\pi_\phi$ is invoked at deployment to map states to actions.
The computational cost of $Q_\theta$ is incurred only during training,
whereas the cost of $\pi_\phi$ is incurred at every decision step
during deployment. This observation is the design pivot of the
proposed method.
\section{LAC: Light Actor, Deep Critic}
\label{sec:method}

In this section, we propose LAC (\textbf{L}ight \textbf{A}ctor, deep
\textbf{C}ritic), an offline actor--critic method that operationalizes
the asymmetric roles of the actor and the critic. Specifically, LAC scales the critic to a
deep network and pairs it with a lightweight deterministic actor, so
that capacity is invested in the network that is discarded after
training and the network that is invoked at every decision step
remains small. The remainder of this section describes why naively
deepening the critic fails in offline RL and how LAC addresses each
failure.

\subsection{Three Failure Modes of Deep Critics}
\label{sec:method_failures}

A natural question raised by this asymmetry is why the offline RL
literature has not already scaled critics deep. The reason is that simply increasing the
depth of an MLP critic destabilizes value-based
learning~\cite{sinha2020d2rl,bjorck2021deeperrl,peng2024deadly},
and the underlying failure is not attributable to a single source. We
identify three distinct failure modes that arise when critics are
scaled in the offline regime, each operating on a different axis and
each requiring a different remedy.

\paragraph{\textbf{(F1) Optimization Failure}}
A plain MLP critic exhibits the standard pathologies of deep
feedforward networks, including vanishing or exploding gradients,
dead activations, and ill-conditioned Hessians. In an online setting,
these can sometimes be alleviated by additional exploration; in the
offline setting, the dataset is fixed and a critic that fails to
optimize cannot recover. As we will show empirically, a plain MLP
critic fails to learn meaningful policies regardless of depth on our
benchmarks.

\paragraph{\textbf{(F2) Bootstrap-noise Amplification}}
A second failure arises specifically from the bootstrap structure of
TD learning. The critic update in \eqref{eq:critic_loss_prelim} uses
the prediction of the critic itself at the next state as part of the target
through \eqref{eq:bellman_target_prelim}. A deeper critic has strictly
more capacity to fit noise, including the noise present in its own
predictions, and the $1$-step bootstrap folds the fitted noise back
into the next target at every iteration. Errors therefore compound
across updates rather than dissipating, and the failure scales with
critic depth. This mechanism is invisible at shallow depths since a
small critic cannot fit enough noise to make the loop dangerous; it
appears only as the critic grows, producing a coupling between depth
and bootstrap horizon.

\paragraph{\textbf{(F3) Value-range Drift}}
A third failure occurs even when bootstrap noise is suppressed. The
MSE regression target in \eqref{eq:critic_loss_prelim} places no
prior on the magnitude of $Q$-values, so the critic is free to output
any real number and a deep critic with substantial representational
capacity can slowly drift its outputs outside any reasonable scale.
Once the critic has drifted far enough, instabilities accumulate
until training enters a divergent regime, producing a characteristic
\emph{mid-training collapse} pattern in which training appears
healthy for hundreds of thousands of steps and then fails abruptly.

\smallskip\noindent
The three mechanisms address distinct failure modes: an architectural
problem (F1), a target-construction problem (F2), and a loss-function
problem (F3). We find empirically that no single ingredient is
sufficient to scale critic depth in the offline regime. LAC
therefore combines one ingredient against each failure, as described
next.

\subsection{LAC Critic: Three Complementary Ingredients}
\label{sec:method_critic}

\paragraph{\textbf{(I1) Residual MLP Backbone, against (F1)}}
We replace the plain MLP critic with a residual MLP (ResMLP) Backbone
with layer
normalization~\cite{nauman2024bro,doro2023bbf,sinha2020d2rl}.
Skip connections restore gradient flow through depth, and layer
normalization stabilizes activations against the magnitude shifts
induced by stacking many layers. With (I1) in place, the optimizer
can train the network end-to-end; whether the trained critic produces
useful predictions is a separate question addressed by (I2) and (I3).

\paragraph{\textbf{(I2) $n$-step Bootstrap Targets, against (F2)}}
We replace the $1$-step TD target in
\eqref{eq:bellman_target_prelim} with an $n$-step
target~\cite{sutton2018reinforcement,kapturowski2019r2d2,hessel2018rainbow},
defined as follows.
\begin{equation}
\label{eq:nstep_target}
\mathcal{T}_n Q_\theta(s_t, a_t) = \sum_{k=0}^{n-1} \gamma^{k} r_{t+k} \,+\, \gamma^{n}\, Q_{\theta^-}\!\big(s_{t+n},\, \pi_{\phi^-}(s_{t+n})\big)
\end{equation}
Compared to \eqref{eq:bellman_target_prelim}, the $n$-step target
incorporates $n$ observed rewards and changes the bootstrap coefficient
from $\gamma$ to $\gamma^n$, thereby reducing reliance on the critic's
own prediction. A noisy deep critic therefore poisons its own targets
less aggressively at each step. Importantly, (I2) is not a generic
performance booster: at shallow critic depths it confers little benefit
because (F2) is dormant, and its effect emerges only as depth grows,
evidencing the coupling between depth and bootstrap noise.

\paragraph{\textbf{(I3) Categorical Cross-Entropy Loss, against (F3)}}
We replace the MSE regression loss with a categorical cross-entropy
loss over a bounded, fixed support $\{z_1, \ldots,
z_I\}$~\cite{bellemare2017c51,hessel2018rainbow}.
The critic outputs a categorical distribution
$p(\cdot \mid s, a)$ over the supports, and its expected value, used
wherever a scalar $Q$ is required, is defined as follows.

\begin{equation}
\label{eq:categorical_expectation}
\mathbb{E}\big[Q_\theta(s_t, a_t)\big] = \sum_{i=1}^{I} z_i \cdot p_i(s_t, a_t \mid \theta)
\end{equation}

The scalar target $\mathcal{T}n Q\theta(s_t,a_t)$ in \eqref{eq:nstep_target} is projected onto the categorical support using a C51-style two-hot interpolation~\cite{bellemare2017c51,hessel2018rainbow} to produce a target distribution $\hat p$, against which the cross-entropy is taken. Two properties of (I3) matter for (F3).
First, predictions are confined to the support range by construction,
so the drift dynamics of \eqref{eq:critic_loss_prelim} cannot push
outputs to arbitrarily large magnitudes. Second, the cross-entropy
loss converts an unbounded regression problem into a bounded
classification problem over a fixed range, which empirically blocks
the abrupt mid-training collapse that MSE exhibits even with $n$-step
in place.

\smallskip\noindent
The three ingredients are complementary, and removing any one
re-introduces its corresponding failure mode. Together, they allow
the critic to be scaled deep.

\subsection{LAC Actor: Lightweight Deterministic}
\label{sec:method_actor}

Given a critic strong enough to provide an accurate learning signal,
the role of the actor reduces to amortized inference of the policy
gradient. We adopt the simplest parametrization available: a
deterministic feedforward MLP $\pi_\phi: \mathcal{S} \to \mathcal{A}$
trained with the standard offline actor objective in
\eqref{eq:actor_bc_prelim}, i.e., the deterministic policy gradient
with a BC regularizer~\cite{fujimoto2021minimalist,tarasov2023rebrac}. We
do not employ any expressive generative parametrization such as
diffusion or flow-matching: producing an action at deployment requires
a single forward pass through a small MLP. We report two
configurations, \textbf{LAC-S} and \textbf{LAC-L}, which differ only
in actor width and depth.\footnote{LAC-S uses a small two-layer actor,
while LAC-L is matched to the size of baselines for
controlled comparison. Concrete architectural details are deferred to Section~\ref{sec:exp_setup} and Appendix~\ref{appendix:hyperparameters}.} Both configurations share
the same ResMLP categorical critic.

A natural concern is whether a lightweight deterministic actor can remain competitive with diffusion- or flow-matching actors on datasets that exhibit multimodal behavior. We address this empirically in Section 5.5, where we show that a deterministic actor remains competitive when paired with the LAC critic, suggesting that a sufficiently strong critic can substantially reduce the performance advantage of actor-side generative expressivity.

\subsection{Critic Architecture}
\label{sec:method_impl}
In this subsection, we provide a more formal description of the LAC critic. 

\paragraph{\textbf{Residual MLP Backbone}}
Let $x_0 \in \mathbb{R}^{d}$ denote the input embedding obtained by
concatenating the state and action features and projecting them
linearly to a hidden dimension $d$. The ResMLP critic is composed of
$B$ residual blocks $g_1, \ldots, g_B$, each defined as follows.
\begin{equation}
\label{eq:resmlp_block}
g_b(x) = x + f_b(x), \quad f_b(x) = \big(\sigma \circ \mathrm{LN} \circ \mathrm{Dense}\big)^{L}(x)
\end{equation}
Herein, $f_b$ is a sequence of $L$ sub-layers, each composed of a
dense projection followed by layer normalization (LN) and a
non-linearity $\sigma(\cdot)$. The skip connection $x + f_b(x)$
ensures that the optimizer sees a path of unit Jacobian through every
block at initialization, which is necessary for stable training of
very deep critics~\cite{wang2025thousand}. Stacking $B$ such
blocks yields the trunk
\begin{equation}
\label{eq:resmlp_trunk}
h(x_0) = (g_B \circ \cdots \circ g_1)(x_0),
\end{equation}
which is followed by a final LayerNorm and a linear head producing
the categorical logits over the support $\{z_1, \ldots, z_I\}$.

\paragraph{\textbf{Categorical Target Projection.}}
Given the $n$-step scalar target
$y = \mathcal{T}_n Q_\theta(s_t,a_t)$ in \eqref{eq:nstep_target},
we project $y$ onto the fixed categorical support using a
C51-style two-hot interpolation~\cite{bellemare2017c51, hessel2018rainbow}.
Let
\[
\bar y = \operatorname{clip}(y, v_{\min}, v_{\max}), \qquad
b = \frac{\bar y-v_{\min}}{\delta_z},
\]
where $\delta_z=(v_{\max}-v_{\min})/(I-1)$, and let
$l=\lfloor b \rfloor$ and $u=\lceil b \rceil$.
The target distribution is constructed as
\begin{equation}
\begin{cases}
\hat p_l = 1, & l=u,\\
\hat p_l = u-b,\quad \hat p_u=b-l, & l\neq u,
\end{cases}
\label{eq:c51_projection}
\end{equation}
with all other entries of $\hat p$ set to zero.
The critic is trained by minimizing the cross-entropy between
$\hat p$ and the predicted distribution $p(\cdot\mid s_t,a_t)$.

\paragraph{\textbf{Target Network Update}}
A target critic $Q_{\theta^-}$ is maintained as a Polyak average of
$Q_\theta$ and is used to compute the next-state value in
\eqref{eq:nstep_target}. The target parameters are updated at every
gradient step as follows.
\begin{equation}
\label{eq:polyak}
\theta^- \leftarrow \tau\, \theta + (1 - \tau)\, \theta^-
\end{equation}
Herein, $\tau \in (0, 1]$ is a small averaging coefficient. An
analogous target $\pi_{\phi^-}$ is maintained for the actor and is
used only to compute the next-state action in the target
\eqref{eq:nstep_target}, following the standard practice of
TD3-style methods~\cite{fujimoto2021minimalist}.

\subsection{Algorithm Summary}
\label{sec:method_summary}

The overall procedure of LAC is summarized in
Algorithm~\ref{alg:lac}. The critic combines (I1) a residual depth,
(I2) $n$-step bootstrap targets, and (I3) a categorical cross-entropy
loss, each addressing a distinct failure mode of deep critics in
offline RL. The actor is a lightweight deterministic MLP trained with
the standard objective in \eqref{eq:actor_bc_prelim}. Inference cost
at deployment is determined by the actor alone, allowing LAC to
achieve the performance of expressive generative actor baselines at
substantially lower per-action latency.

\section{Simulation Results}
\label{sec:experiments}
We empirically validate three claims that follow from the design
principle articulated in Section~\ref{sec:introduction}:
(i) LAC matches or exceeds generative actor baselines on standard
offline RL benchmarks while reducing per-action inference cost by up to $4\times$\rev{ over multi-step generative actors};
(ii) critic-side capacity, when properly stabilized, is the productive
axis for offline RL scaling, whereas actor-side capacity saturates
rapidly\rev{ on most environments}; and
(iii) the critic-side ingredients of LAC act as a drop-in improvement
that benefits any actor parametrization, including diffusion- and
flow-matching actors.

\subsection{Simulation Setup}
\label{sec:exp_setup}

\paragraph{\textbf{Benchmarks}}
We follow the experimental protocol of recent flow-based offline RL
work~\cite{park2025fql} for direct comparability, and evaluate on 
$7$ challenging environments from OGBench~\cite{park2025ogbench}, 
a recent suite designed to stress-test offline RL methods on 
long-horizon and high-dimensional continuous control. The 
environments span three categories. 
\begin{itemize}
    \item \textbf{Ant navigation ($8$-DoF):} \texttt{ant-large} 
    and \texttt{ant-giant}, where a quadruped robot must navigate 
    increasingly large maze layouts to a goal location.
    \item \textbf{Humanoid navigation ($21$-DoF):} 
    \texttt{hum-medium} and \texttt{hum- large}, 
    where a whole-body humanoid solves the same maze layouts under 
    a much higher-dimensional action space.
    \item \textbf{Manipulation:} \texttt{scene} requires a sequence 
    of pick-place and tool-use sub-actions, while \texttt{puzzle-3x3} 
    and \texttt{puzzle-4x4} require solving sliding-tile puzzles 
    that combine continuous low-level control with discrete 
    combinatorial task structure.
\end{itemize}
Each environment provides $5$ evaluation goals 
(\texttt{task1}--\texttt{task5}), yielding \rev{$35$} tasks in total. 
For ablation and analysis experiments 
(Sections~\ref{sec:exp_critic_depth}--\ref{sec:exp_qcal}), we report 
results on the \texttt{task1} variant of each environment unless 
stated otherwise. Further details on the environments are provided 
in Appendix~\ref{appendix:benchmark}.

\paragraph{\textbf{Baselines}}
We compare against $11$ offline RL methods across three actor families.
\begin{itemize}
    \item \textbf{Gaussian / deterministic.}
    BC, IQL~\cite{kostrikov2022iql},
    ReBRAC~\cite{tarasov2023rebrac}, and
    TD3+BC~\cite{fujimoto2021minimalist}, the closest reference 
    to our deterministic actor configuration.
    \item \textbf{Diffusion.}
    IDQL~\cite{hansen2023idql}, SRPO~\cite{chen2024srpo}, and
    CAC~\cite{ding2024cac}. SRPO and CAC retain the expressivity 
    of diffusion-based actors while reducing inference cost through 
    single-step parametrizations.
    \item \textbf{Flow.}
    FAWAC, FBRAC, IFQL, and FQL~\cite{park2025fql}. FQL similarly 
    distills a flow-matching actor into a single forward pass to 
    accelerate inference.
\end{itemize}

\paragraph{\textbf{Metrics and Implementation}}
We report binary success rate (\%) on all OGBench tasks, averaged over
$4$ random seeds, following the OGBench evaluation
protocol~\cite{park2025ogbench}.
Inference latency is measured as wall-clock time per action
prediction on a single NVIDIA RTX 3090 with batch size $1$. LAC uses a $32$-layer ResMLP
critic \rev{($\approx 2.15$M parameters)} with $51$-atom categorical head and $n = 4$ bootstrap targets;
\textbf{LAC-S} uses a $[256, 256]$ actor ($\approx 0.13$M parameters)
and \textbf{LAC-L} uses a $[512, 512, 512, 512]$ actor matched in
size to our flow-based baselines. Full hyperparameters are listed in
Appendix~\ref{appendix:hyperparameters}.

\begin{algorithm}[t]
\caption{\textbf{LAC: Light Actor, Deep Critic}}
\label{alg:lac}
\begin{algorithmic}[1]
\STATE \textbf{Input:} Dataset $\mathcal{D}$ with $n$-step transitions
$(s_t, a_t, s_{t+n}, G_t^{(n)})$, deep ResMLP critic $Q_\theta$ with
categorical head over support $\{z_i\}_{i=1}^{I}$, target critic
$Q_{\theta^-}$, MLP actor $\pi_\phi$ \rev{with target $\pi_{\phi^-}$}, $n$-step horizon $n$
\STATE \textbf{Hyperparameters:} Learning rates $\alpha_Q, \alpha_\pi$,
target update rate $\tau$, total \rev{gradient steps $T$}
\FOR{\rev{$\mathrm{step} = 1$ to $T$}}
    \STATE Sample mini-batch $B$ from $\mathcal{D}$
    \STATE Compute next-state value as the expectation of the
           categorical critic: \\
            $q_{\text{next}} = \sum_{i=1}^{I} z_i \cdot
            p_i\!\big(s_{t+n},\, \rev{\pi_{\phi^-}}(s_{t+n}) \mid \theta^-\big)$
    \STATE Compute scalar target:
           $y = G_t^{(n)} + \gamma^{n} \cdot q_{\text{next}}$
    \STATE Project $y$ onto the categorical support via a C51-style two-hot interpolation to obtain the target distribution $\hat p$ (see \eqref{eq:c51_projection}).
    \STATE Update critic:
           $\theta \leftarrow \theta - \alpha_Q\, \nabla_\theta
           \mathrm{CE}\big(\hat{p},\, Q_\theta(s_t, a_t)\big)$
    \STATE Update actor:
           $\phi \leftarrow \phi - \alpha_\pi\, \nabla_\phi
           \mathcal{L}(\phi)$ using \eqref{eq:actor_bc_prelim}
    \STATE Soft-update target\rev{s}:
           $\theta^- \leftarrow \tau \theta + (1 - \tau) \theta^-$\rev{ and
           $\phi^- \leftarrow \tau \phi + (1 - \tau) \phi^-$}
\ENDFOR
\end{algorithmic}
\end{algorithm}

\begin{table*}[t]
\centering
\caption{Success rates on $7$ OGBench environments ($5$ task variants averaged). The best result in each row is shown in \textbf{bold}, and the second-best is \underline{underlined}.}
\label{tab:main_results}
\renewcommand{\arraystretch}{0.85}
\setlength{\tabcolsep}{3pt}
\scriptsize
\resizebox{\textwidth}{!}{%
\begin{tabular}{l|cccc|ccc|cccc|cc}
\toprule
& \multicolumn{4}{c|}{\textbf{Gaussian / Det.}} & \multicolumn{3}{c|}{\textbf{Diffusion}} & \multicolumn{4}{c|}{\textbf{Flow}} & \multicolumn{2}{c}{\textbf{Ours}} \\
\cmidrule(lr){2-5} \cmidrule(lr){6-8} \cmidrule(lr){9-12} \cmidrule(lr){13-14}
\textbf{Environment} & BC & TD3+BC & IQL & ReBRAC & IDQL & SRPO & CAC & FAWAC & FBRAC & IFQL & FQL & \textbf{LAC-S} & \textbf{LAC-L} \\
\midrule
\texttt{ant-large} & 11{\tiny$\pm$1} & 82{\tiny$\pm$6} & 53{\tiny$\pm$3} & 81{\tiny$\pm$5} & 21{\tiny$\pm$5} & 11{\tiny$\pm$4} & 33{\tiny$\pm$4} & 6{\tiny$\pm$1} & 60{\tiny$\pm$6} & 28{\tiny$\pm$5} & 79{\tiny$\pm$3} & \underline{92{\tiny$\pm$5}} & \textbf{96{\tiny$\pm$1}} \\
\texttt{ant-giant} & 0{\tiny$\pm$0} & 3{\tiny$\pm$4} & 4{\tiny$\pm$1} & 26{\tiny$\pm$8} & 0{\tiny$\pm$0} & 0{\tiny$\pm$0} & 0{\tiny$\pm$0} & 0{\tiny$\pm$0} & 4{\tiny$\pm$4} & 3{\tiny$\pm$2} & 9{\tiny$\pm$6} & \underline{35{\tiny$\pm$23}} & \textbf{79{\tiny$\pm$13}} \\
\texttt{hum-medium} & 2{\tiny$\pm$1} & 12{\tiny$\pm$9} & 33{\tiny$\pm$2} & 22{\tiny$\pm$8} & 1{\tiny$\pm$0} & 1{\tiny$\pm$1} & 53{\tiny$\pm$8} & 19{\tiny$\pm$1} & 38{\tiny$\pm$5} & 60{\tiny$\pm$14} & 58{\tiny$\pm$5} & \textbf{79{\tiny$\pm$10}} & \underline{76{\tiny$\pm$38}} \\
\texttt{hum-large} & 1{\tiny$\pm$0} & 8{\tiny$\pm$7} & 2{\tiny$\pm$1} & 2{\tiny$\pm$1} & 1{\tiny$\pm$0} & 0{\tiny$\pm$0} & 0{\tiny$\pm$0} & 0{\tiny$\pm$0} & 2{\tiny$\pm$0} & 11{\tiny$\pm$2} & 4{\tiny$\pm$2} & \underline{50{\tiny$\pm$18}} & \textbf{69{\tiny$\pm$15}} \\
\texttt{scene} & 5{\tiny$\pm$1} & 0{\tiny$\pm$0} & 28{\tiny$\pm$1} & 41{\tiny$\pm$3} & \underline{46{\tiny$\pm$3}} & 20{\tiny$\pm$1} & 40{\tiny$\pm$7} & 30{\tiny$\pm$3} & 45{\tiny$\pm$5} & 30{\tiny$\pm$3} & \textbf{56{\tiny$\pm$2}} & 29{\tiny$\pm$37} & 41{\tiny$\pm$39} \\
\texttt{puzzle-3x3} & 2{\tiny$\pm$0} & 7{\tiny$\pm$13} & 9{\tiny$\pm$1} & 21{\tiny$\pm$1} & 10{\tiny$\pm$2} & 18{\tiny$\pm$1} & 19{\tiny$\pm$0} & 6{\tiny$\pm$2} & 14{\tiny$\pm$4} & 19{\tiny$\pm$1} & 30{\tiny$\pm$1} & \underline{62{\tiny$\pm$17}} & \textbf{95{\tiny$\pm$1}} \\
\texttt{puzzle-4x4} & 0{\tiny$\pm$0} & 2{\tiny$\pm$3} & 7{\tiny$\pm$1} & 14{\tiny$\pm$1} & \underline{29{\tiny$\pm$3}} & 10{\tiny$\pm$3} & 15{\tiny$\pm$3} & 1{\tiny$\pm$0} & 13{\tiny$\pm$1} & 25{\tiny$\pm$5} & 17{\tiny$\pm$2} & 15{\tiny$\pm$9} & \textbf{32{\tiny$\pm$19}} \\
\midrule
\rowcolor{gray!20}
\textbf{Avg} & 3 & 16 & 19 & 30 & 15 & 9 & 23 & 9 & 25 & 25 & 36 & \underline{52} & \textbf{70} \\
\bottomrule
\end{tabular}%
}
\end{table*}

\subsection{Performance Comparison}

Table~\ref{tab:main_results} reports aggregated success rates over
the $5$ task variants of each environment.\footnote{Baseline numbers 
are taken from~\cite{park2025fql}, except for TD3+BC, which we 
re-run under our setup.} LAC-L achieves the highest average score 
across the suite, and LAC-S recovers most of that performance with 
an actor roughly one-tenth the size. LAC-S alone exceeds all 
generative actor baselines on the \rev{long-horizon locomotion 
environments}. The humanoid suite is the most 
informative comparison point: its $21$-dimensional action space 
yields substantially richer multimodal behavior, which is where 
generative parametrizations are most 
beneficial~\cite{park2025fql}---and indeed IFQL achieves the 
strongest baseline scores on \texttt{hum-medium} and 
\rev{\texttt{hum-large}}. Despite operating with a deterministic actor, LAC matches or surpasses IFQL without requiring multiple integration steps, indicating that a strong critic can substantially narrow the performance gap associated with using a simpler actor.
\rev{LAC-L and LAC-S differ only in actor architecture. Their scores
are close on several environments, but clear gaps remain on
\texttt{ant-giant} and \texttt{puzzle-3x3}
(Section~\ref{sec:exp_capacity}).}
\subsection{Capacity Asymmetry Analysis}
\label{sec:exp_capacity}

We analyze where capacity matters in offline actor--critic methods,
testing the hypothesis that critic capacity, when stabilized, is the
binding constraint while actor capacity saturates rapidly. We study
three axes: scaling the critic, scaling the actor, and varying the
actor parametrization with a fixed deep critic.

\subsubsection{\textbf{Critic Capacity is Productive When Stabilized}}
\label{sec:exp_critic_depth}

\begin{figure}[t]
    \centering
    \includegraphics[width=0.8\columnwidth]{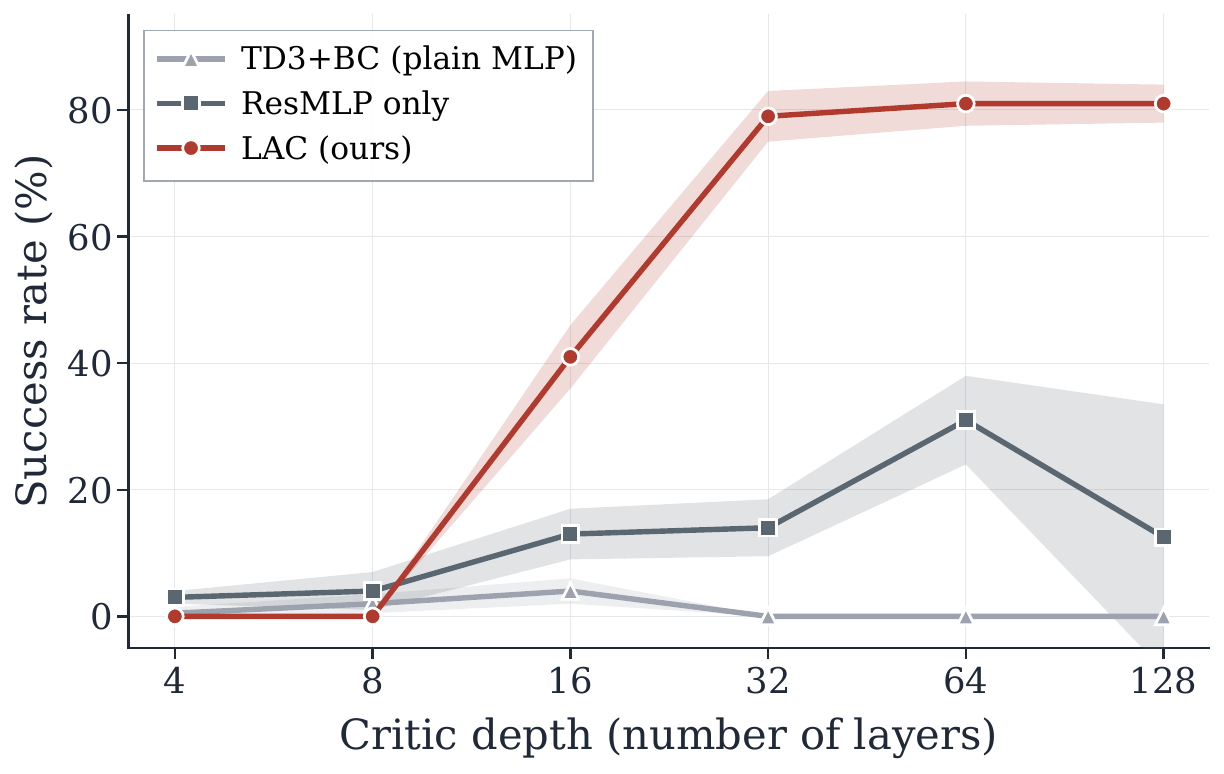}
    \caption{Critic depth scaling on \texttt{hum-large}. LAC scales
    stably with depth while baselines plateau or fail to learn.}
    \label{fig:critic_depth}
\end{figure}

Figure~\ref{fig:critic_depth} plots success rate against critic depth
for three configurations: (i) a TD3+BC-style plain MLP critic with
$1$-step MSE regression; (ii) the same critic with a residual MLP
backbone and LayerNorm but still using $1$-step MSE regression; and
(iii) the full LAC critic recipe with $n$-step bootstrap and
categorical cross-entropy.
\begin{figure}[t]
    \centering
    \includegraphics[width=\columnwidth]{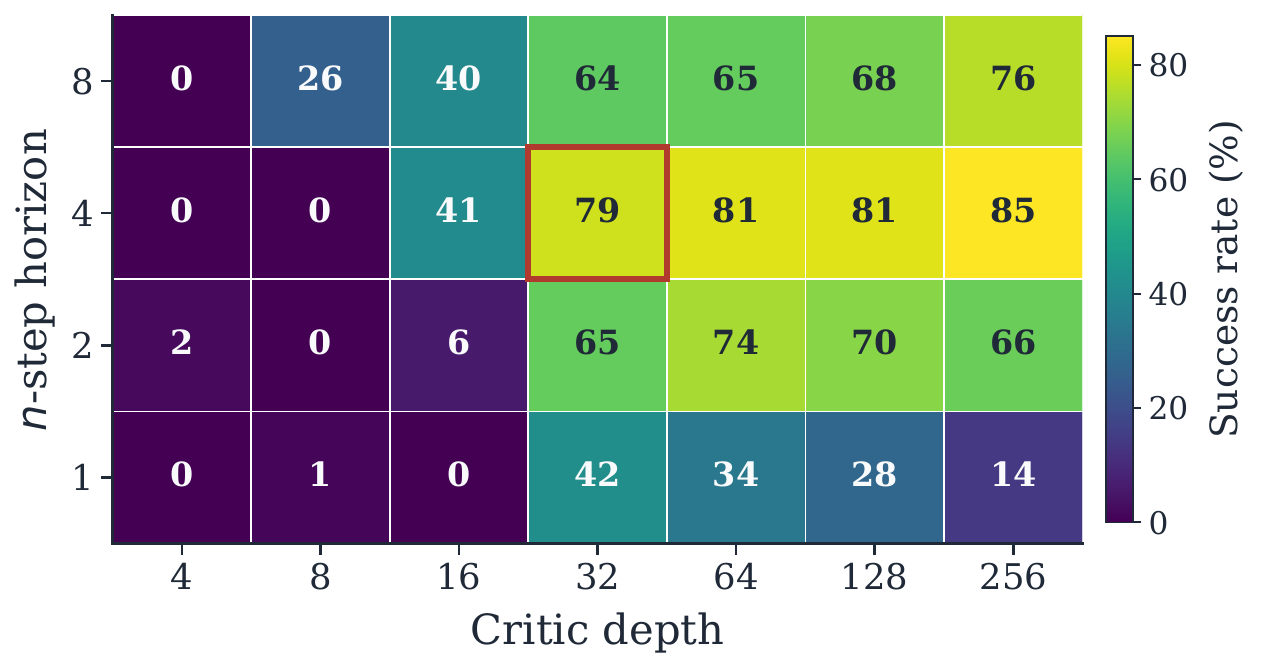}
    \caption{Success rate over critic depth $\times$ $n$-step horizon
    on \texttt{hum-large}, with residual backbone and categorical loss
    fixed. All horizons fail at shallow depth; beyond depth $16$, only
    $n = 1$ degrades with further depth, while $n \ge 2$ remains stable
    and $n = 4$ is consistently strongest.}
    \label{fig:nstep_depth_heatmap}
\end{figure}

The plain MLP baseline fails to learn meaningful policies across the
entire $4$--$128$ layer sweep, consistent with prior reports that
naive depth scaling in offline RL leads to gradient propagation
failures and value
divergence~\cite{sinha2020d2rl,peng2024deadly}. The residual-only backbone reaches roughly $20\%$ but does not improve with further depth and exhibits growing seed variance at depth 128, indicating that gradient flow alone is insufficient.
The full recipe is also near zero at
$4$--$8$ layers, where a critic this small cannot represent the
value landscape and the bounded categorical head constrains a
capacity it does not yet have, but it is the only configuration that
converts depth into performance, rising to $\sim 80\%$ at depth $32$
and saturating thereafter with small seed variance.

\subsubsection{\textbf{$n$-Step Bootstrap is an Enabling Condition for Depth}}
\label{sec:exp_nstep_depth}

To isolate the role of $n$-step bootstrap, we sweep depth and $n$
jointly with the residual backbone and categorical loss held fixed
(Figure~\ref{fig:nstep_depth_heatmap}).

At shallow depths ($\le 8$ layers), all horizons fail: a critic this
small cannot learn the value landscape, so the choice of $n$ is
immaterial. The rows separate from depth $16$ onward, and $n = 1$ is
the only configuration that degrades with additional depth, peaking
at $42\%$ at depth $32$ and decaying to $14\%$ at depth $256$, while
every $n \ge 2$ configuration remains stable. This evidences the
bootstrap-noise mechanism: a $1$-step target folds more of the
critic's own fitted noise back into each subsequent target as depth
grows, and the $\gamma^n$ attenuation breaks the loop. The benefit
is not monotone in $n$, however, as $n = 4$ dominates beyond depth 16
while $n = 8$ falls back to the level of $n = 2$. Depth and
$n$-step are therefore coupled by the bootstrap mechanism, not
independent design choices.

\subsubsection{\textbf{$n$-Step Alone Is Not Enough: Categorical Targets Prevent Mid-Training Collapse}}
\label{sec:exp_nstep_collapse}
\begin{figure}[t]
    \centering
    \includegraphics[width=\columnwidth]{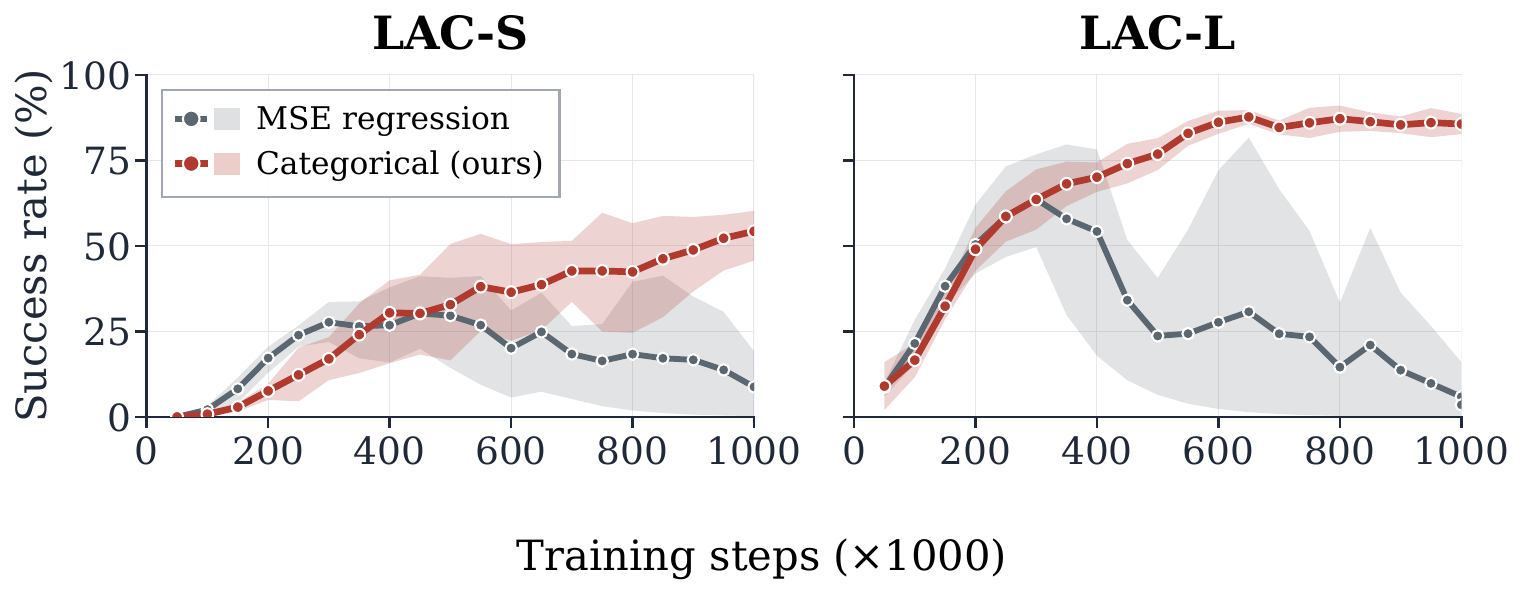}
    \caption{Training curves on \texttt{hum-large} with a $32$-layer
    residual critic and $n = 4$, varying only the critic loss. The
    MSE-regression critic collapses mid-training; the categorical
    critic of LAC remains stable. Shaded: $\pm 1$ std over $4$ seeds.}
    \label{fig:training_collapse}
\end{figure}

Figure~\ref{fig:training_collapse} compares training curves when only
the critic loss is varied, with the residual backbone and $n = 4$
held fixed.
Both configurations rise along essentially identical trajectories for
the first $\sim 300$K steps. The MSE critic then collapses abruptly,
with success rate dropping to near zero over roughly $100$K steps
and seed variance widening dramatically. The categorical critic
exhibits no such collapse. MSE regression places no prior on the
value range, so a deep critic drifts onto unbounded targets;
instabilities accumulate until training enters a divergent regime.
The categorical loss bounds the targets the bootstrap loop can
generate, complementing the variance reduction of $n$-step.

\subsubsection{\textbf{Actor Capacity Saturates Rapidly}}
\label{sec:exp_actor_scaling}
\begin{figure}[t]
    \centering
    \includegraphics[width=\columnwidth]{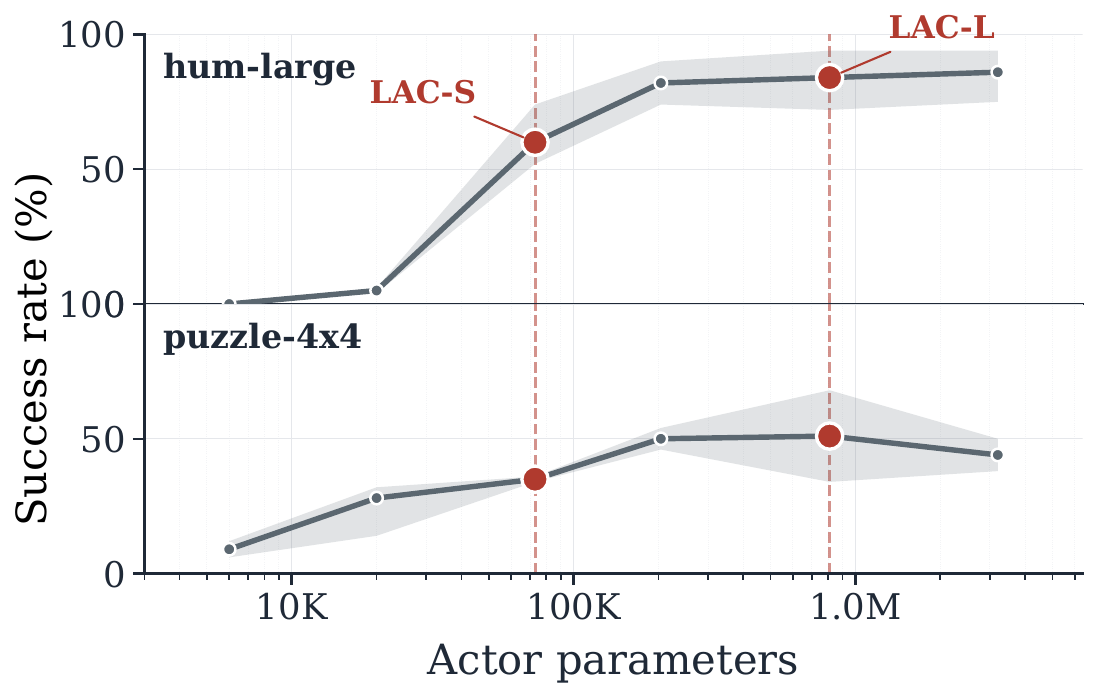}
    \caption{Actor scaling on \texttt{hum-large} (top) and
    \texttt{puzzle-4x4} (bottom) with the deep critic fixed.
    Performance saturates at modest actor sizes; LAC-S and LAC-L
    bracket the plateau.}
    \label{fig:actor_scaling}
\end{figure}

Figure~\ref{fig:actor_scaling} shows how performance varies with
actor parameter count, holding the critic fixed to the $32$-layer
ResMLP configuration. Performance rises sharply up to roughly
$10^{5}$ parameters and plateaus thereafter; LAC-S and LAC-L bracket
the plateau.
This is the empirical counterpart of the capacity asymmetry: with the
critic strong, the role of the actor reduces to amortized inference of the
policy gradient, and additional actor capacity \rev{brings little
further benefit on these environments}. Capacity invested in the actor incurs inference cost at every
decision step \rev{for rapidly diminishing returns}; the same capacity invested
in the critic incurs cost only during training.

\subsection{Ablation Studies and Mechanistic Analysis}
\label{sec:exp_ablation}

We isolate which critic-side ingredients contribute and how they
improve the learning signal received by the actor.

\subsubsection{\textbf{Component Ablation}}
\label{sec:exp_component}

Table~\ref{tab:component_ablation} reports the effect of removing
each critic-side ingredient individually while holding the rest
fixed.
\begin{table}[t]
\centering
\caption{Component ablation \rev{on the \texttt{task1} variant of
each environment}. Each row uses a $32$-layer residual
critic with different combinations of $n$-step bootstrap and
categorical loss.}
\label{tab:component_ablation}
\renewcommand{\arraystretch}{1.1}
\setlength{\tabcolsep}{6pt}
\small
\begin{tabular}{l|ccc|c}
\toprule
\textbf{Configuration} & \textbf{H-large} & \textbf{A-giant} & \textbf{P-4x4} & \textbf{Avg} \\
\midrule
LAC ($n$-step + categorical)        & $\textbf{79} \pm \textbf{6}$  & $\textbf{86} \pm \textbf{3}$ & $\textbf{55} \pm \textbf{6}$ & \textbf{73} \\
\midrule
$-$ $n$-step (use $1$-step TD)      & $42 \pm 3$  & $38 \pm 15$ & $18 \pm 2$ & 33 \\
$-$ Categorical (use MSE)           & $7 \pm 12$  & $72 \pm 12$ & $32 \pm 15$ & 37 \\
$-$ Both ($1$-step TD + MSE)        & $15 \pm 5$  & $1 \pm 1$ & $12 \pm 3$ & 9 \\
\bottomrule
\end{tabular}
\end{table}
The $n$-step and categorical ingredients exhibit the asymmetry 
expected from the analyses above: removing $n$-step produces 
depth-dependent degradation consistent with 
Figure~\ref{fig:nstep_depth_heatmap}, while removing the categorical 
loss produces the mid-training collapse of 
Figure~\ref{fig:training_collapse}. The impact of each ingredient 
also varies with task properties. On \texttt{ant-giant}, the long 
horizon makes $n$-step particularly valuable independently of critic 
depth, so removing it causes the most severe drop ($86 \to 38$). 
The categorical loss, in contrast, matters most when the action 
space is high-dimensional: on \texttt{hum-large} ($21$-DoF) its 
removal causes near-total collapse ($79 \to 7$), whereas on the 
lower-dimensional \texttt{ant-giant} ($86 \to 72$) and 
\texttt{puzzle-4x4} ($55 \to 32$) the degradation is substantially 
milder. The two ingredients address distinct 
failure modes---bootstrap-noise amplification and value-range 
drift---and both are required to scale critic depth.

\subsubsection{\textbf{Spatial $Q$-Value Resolution}}
\label{sec:exp_qcal}

\begin{figure}[t]
    \centering
    \includegraphics[width=\columnwidth]{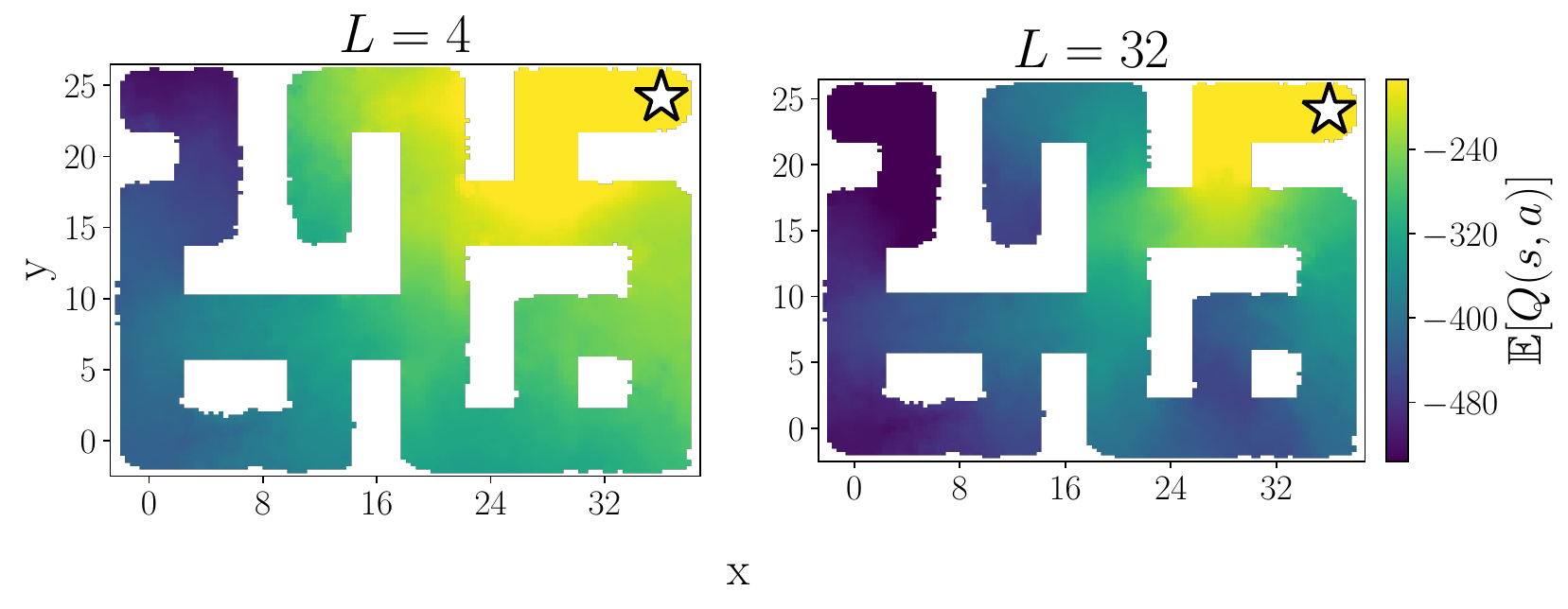}
\caption{Spatial value landscape on \texttt{hum-large}. For each 
spatial location $(x, y)$, we average the predicted $Q$-value over 
actions sampled at that location. The shallow $4$-layer critic 
(left) yields a blurred landscape, whereas the $32$-layer LAC 
critic (right) produces a sharper gradient aligned with the 
distance to the goal (star).}
    \label{fig:q_cal}
\end{figure}

Figure~\ref{fig:q_cal} provides a qualitative view of why critic 
depth matters. For each dataset state on the \texttt{hum-large} 
maze, we evaluate the trained critic and average the predicted 
$Q$-value over actions sampled at that state, producing a $2$D 
heatmap over the spatial coordinates of the agent. This visualizes 
the value landscape the critic has learned, which is the quantity 
the actor relies on to extract goal-directed behavior. The shallow $4$-layer critic produces a blurred landscape in which 
states at markedly different distances from the goal receive 
similar values, leaving the actor without a clear gradient to 
follow. The deep $32$-layer LAC critic, by contrast, assigns 
visibly distinct values across the map, with a smooth gradient 
that aligns with the geometric distance to the goal. The lightweight deterministic actor benefits directly from this spatial resolution, suggesting that a sufficiently informative critic can support strong policy learning without requiring the actor to explicitly model a multimodal action distribution.

\subsection{Inference Efficiency and Drop-in Compatibility}
\label{sec:exp_efficiency}

Table~\ref{tab:latency} reports per-action wall-clock latency for
all methods under identical hardware and measurement protocol.
\begin{table}[t]
\centering
\caption{Per-action inference latency on a single NVIDIA RTX 3090,
batch size $1$. Speedup is relative to LAC-S.}
\label{tab:latency}
\renewcommand{\arraystretch}{1.0}
\setlength{\tabcolsep}{6pt}
\small
\begin{tabular}{l|ccc}
\toprule
\textbf{Method} & \textbf{Actor (inference)} & \textbf{Latency (ms)} & \textbf{Speedup} \\
\midrule
IQL                       & Gaussian, 1 step              & 0.30  & $1.25\times$ \\
ReBRAC                    & Deterministic, 1 step         & 0.27  & $1.13\times$ \\
IDQL                      & Diffusion, 10 steps           & 0.95  & $3.96\times$ \\
SRPO                      & Diffusion, 1 step             & 0.28  & $1.17\times$ \\
CAC                       & Distilled Diffusion, 1 step   & 0.31  & $1.29\times$ \\
FBRAC                     & Flow, 10 steps                & 0.55  & $2.29\times$ \\
FAWAC                     & Flow, 10 steps                & 0.56  & $2.33\times$ \\  
IFQL                      & Flow, 10 steps                & 0.64  & $2.67\times$ \\  
FQL                       & Distilled Flow, 1 step        & 0.27  & $1.13\times$ \\
\midrule
LAC-L                     & Det.\ $[512]\times 4$         & 0.31  & $1.29\times$ \\
LAC-S                     & Det.\ $[256]\times 2$         & \textbf{0.24} & $1.00\times$ \\
\bottomrule
\end{tabular}
\end{table}
Generative actors incur per-action costs that scale with the number
of denoising or integration steps: IDQL, FBRAC, FAWAC, and IFQL require
multiple iterative steps at inference and are $2\text{--}4\times$
slower than single-step baselines. FQL eliminates this cost via
one-step distillation. Both configurations of LAC produce each action
in a single forward pass: LAC-S achieves the fastest per-action latency at $0.24$ms, matching or beating all single-step baselines, while LAC-L incurs only
a modest overhead at $0.31$ms despite using a larger actor. \rev{The
$4\times$ speedup is thus specific to multi-step actors, and the
advantage over the one-step distilled baselines (FQL, SRPO, CAC) is
marginal.} Combined
with the performance results, LAC-S occupies the top-left of the
performance-latency plane.

To further demonstrate that the inference efficiency of LAC does not
come at the cost of compatibility with existing methods,
Table~\ref{tab:actor_arch} shows the effect of replacing the critic
of each baseline with the LAC critic, while keeping its original
actor parametrization and policy-extraction objective intact.

\begin{table}[t]
\centering
\caption{Replacing the critic of various offline RL methods with
the LAC critic, while keeping their actor parametrization and
policy-extraction objective intact. \rev{Success rates ($\%$) on the
\texttt{task1} variant of \texttt{hum-medium}.}}
\label{tab:actor_arch}
\renewcommand{\arraystretch}{1.1}
\setlength{\tabcolsep}{8pt}
\small
\begin{tabular}{l|cc|c}
\toprule
\textbf{Method} & \textbf{Baseline} & \textbf{+ LAC critic} & \textbf{Gain} \\
\midrule
TD3+BC (Deterministic)        & 10 & 46 & \textbf{+36} \\
FAWAC (flow-matching)         & 6 & 30 & \textbf{+24} \\
FQL (Distilled Flow)          & 19 & 57 & \textbf{+38} \\
IDQL (Diffusion)              & 1  & 31 & \textbf{+30} \\
CAC (Distilled Diffusion)     & 38 & 97 & \textbf{+59} \\
\bottomrule
\end{tabular}
\end{table}

Across all actor families, replacing the critic with the LAC critic
\rev{yields consistent gains of $+24$ to $+59$ points under the
identical protocol}, indicating that the critic-side
ingredients of LAC---residual backbone, $n$-step targets, categorical
loss---function as a drop-in enhancement rather than a single-point
algorithm. \rev{We regard this transferability as the central
empirical result of this work. Actor parametrizations are
nonetheless not interchangeable even under the LAC critic. The
distilled-diffusion actor of CAC reaches $97\%$, so the
deterministic actor of LAC is best read as the most
inference-efficient point on this spectrum rather than as evidence
that actor expressivity is redundant.} The implication is that the
inference savings of LAC are not paid for by an algorithm
restriction: the same critic recipe accelerates and improves a
broad family of offline RL methods regardless of their actor
parametrization.

\section{Conclusion}
\label{sec:conclusion}
We proposed LAC, an offline actor--critic method that 
operationalizes a simple observation: the critic shapes the actor 
during training but is discarded at deployment. Investing capacity 
in the critic therefore pays a one-time training cost, whereas 
investing capacity in the actor pays a recurring inference cost. 
Realizing this design in offline RL requires resolving three 
distinct challenges that arise when critics are scaled deep: 
optimization, bootstrap-noise amplification, and value-range drift, 
which LAC addresses with a residual MLP backbone, $n$-step 
bootstrap targets, and a categorical cross-entropy loss, 
respectively. Paired with a lightweight deterministic actor, LAC 
matches the strongest diffusion- and flow-matching baselines on 
OGBench while reducing per-action inference latency by up to 
\rev{$4\times$ over multi-step generative actors while matching
one-step distilled policies without a distillation stage. The
critic recipe transfers as a drop-in improvement to other actor
parametrizations, which we view as the central empirical result.}
Asymmetric capacity 
allocation offers a complementary direction to scaling generative 
actors.

\paragraph{\textbf{Limitation}}
The present evaluation is restricted to state-based OGBench
environments, in line with the protocol of recent offline RL
work~\cite{park2025fql,park2025ogbench}, and to simulated
benchmarks rather than real-robot deployment. Extending the
analysis to pixel-based observations and to real-robot settings
remains an interesting direction for future work.


\makeatletter
\@ifundefined{isChecklistMainFile}{
  \newif\ifreproStandalone
  \reproStandalonetrue
}{
  \newif\ifreproStandalone
  \reproStandalonefalse
}
\makeatother

\begin{appendices}
\begin{table}[h]
\centering

\caption{LAC hyperparameters used across all OGBench environments.}
\label{tab:hyperparams}
\renewcommand{\arraystretch}{1.15}
\setlength{\tabcolsep}{14pt}
\small
\begin{tabular}{ll}
\toprule
\textbf{Hyperparameter} & \textbf{Value} \\
\midrule
\rowcolor{gray!20}\multicolumn{2}{c}{\textit{Critic (ResMLP with categorical head)}} \\
\midrule
Embedding dimension                   & $256$ \\
Number of residual blocks             & $8$ \\
Sub-layers per block                  & $4$ \\
Total dense layers                    & $32$ \\
Activation                            & ReLU~\cite{nair2010relu} \\
Normalization                         & LayerNorm \\
Categorical atoms ($I$)               & $51$ \\
Support range $[v_{\min}, v_{\max}]$  & $[-1/(1-\gamma),\, 0]$ \\
Critic ensembles                      & $1$ \\
\midrule
\rowcolor{gray!20}\multicolumn{2}{c}{\textit{Actor (deterministic MLP)}} \\
\midrule
LAC-S width $\times$ depth            & $[256] \times 2$ \\
LAC-L width $\times$ depth            & $[512] \times 4$ \\
Activation                            & GELU~\cite{hendrycks2016gelu} \\
Output activation                     & $\tanh$ \\
\midrule
\rowcolor{gray!20}\multicolumn{2}{c}{\textit{Optimization}} \\
\midrule
Optimizer                             & Adam~\cite{kingma2014adam} \\
Learning rate (critic and actor)      & $3 \times 10^{-4}$ \\
Batch size                            & $256$ \\
Polyak rate $\tau$                    & $5 \times 10^{-3}$ \\
Discount factor $\gamma$              & $0.99$ / $0.995$ / $0.999$\textsuperscript{$\dagger$} \\
Gradient steps                        & $1$M \\
\midrule
\rowcolor{gray!20}\multicolumn{2}{c}{\textit{Bootstrap and BC}} \\
\midrule
$n$-step horizon                      & $4$ \\
BC coefficient $\lambda$              & $10$ / $20$ / $300$ / $1000$\textsuperscript{$\ddagger$} \\
\bottomrule
\multicolumn{2}{l}{\footnotesize\textsuperscript{$\dagger$}\,$0.99$ for \texttt{scene}, $0.995$ for \texttt{ant-*}/ \texttt{puzzle-*}, $0.999$ for \texttt{hum-*}.} \\
\multicolumn{2}{l}{\footnotesize\textsuperscript{$\ddagger$}\,$10$ for \texttt{ant-*}, $20$ for \texttt{hum-*}, $300$ for \texttt{scene}, $1000$ for \texttt{puzzle-*}.} \\
\end{tabular}
\end{table}




\begin{table*}[h]
\centering
\caption{Full OGBench success rates ($\%$) on all \rev{$35$} tasks. The best result in each row is shown in \textbf{bold}, and the second-best is \underline{underlined}.}
\label{tab:ogbench_full}
\renewcommand{\arraystretch}{0.45}
\setlength{\tabcolsep}{2pt}
\tiny
\resizebox{0.8\textwidth}{!}{%
\begin{tabular}{ll|cccc|ccc|cccc|cc}
\toprule
& & \multicolumn{4}{c|}{\textbf{Gaussian / Det.}} & \multicolumn{3}{c|}{\textbf{Diffusion}} & \multicolumn{4}{c|}{\textbf{Flow}} & \multicolumn{2}{c}{\textbf{Ours}} \\
\cmidrule(lr){3-6} \cmidrule(lr){7-9} \cmidrule(lr){10-13} \cmidrule(lr){14-15}
\textbf{Env} & \textbf{Task} & BC & TD3+BC & IQL & ReBRAC & IDQL & SRPO & CAC & FAWAC & FBRAC & IFQL & FQL & \textbf{L-S} & \textbf{L-L} \\
\midrule
\multirow{5}{*}{\texttt{ant-l}}
& 1 & 0{\tiny$\pm$0} & 86{\tiny$\pm$2} & 48{\tiny$\pm$9} & 91{\tiny$\pm$10} & 0{\tiny$\pm$0} & 0{\tiny$\pm$0} & 42{\tiny$\pm$7} & 1{\tiny$\pm$1} & 70{\tiny$\pm$20} & 24{\tiny$\pm$17} & 80{\tiny$\pm$8} & \underline{92{\tiny$\pm$4}} & \textbf{96{\tiny$\pm$0}} \\
& 2 & 6{\tiny$\pm$3} & 77{\tiny$\pm$3} & 42{\tiny$\pm$6} & \underline{88{\tiny$\pm$4}} & 14{\tiny$\pm$8} & 4{\tiny$\pm$4} & 1{\tiny$\pm$1} & 0{\tiny$\pm$1} & 35{\tiny$\pm$12} & 8{\tiny$\pm$3} & 57{\tiny$\pm$10} & 83{\tiny$\pm$3} & \textbf{94{\tiny$\pm$2}} \\
& 3 & 29{\tiny$\pm$5} & 92{\tiny$\pm$0} & 72{\tiny$\pm$7} & 51{\tiny$\pm$18} & 26{\tiny$\pm$8} & 3{\tiny$\pm$2} & 49{\tiny$\pm$10} & 12{\tiny$\pm$4} & 83{\tiny$\pm$15} & 52{\tiny$\pm$17} & 93{\tiny$\pm$3} & \underline{97{\tiny$\pm$2}} & \textbf{98{\tiny$\pm$2}} \\
& 4 & 8{\tiny$\pm$3} & 79{\tiny$\pm$3} & 51{\tiny$\pm$9} & 84{\tiny$\pm$7} & 62{\tiny$\pm$25} & 45{\tiny$\pm$19} & 17{\tiny$\pm$6} & 10{\tiny$\pm$3} & 37{\tiny$\pm$18} & 18{\tiny$\pm$8} & 80{\tiny$\pm$4} & \underline{91{\tiny$\pm$3}} & \textbf{95{\tiny$\pm$0}} \\
& 5 & 10{\tiny$\pm$3} & 78{\tiny$\pm$3} & 54{\tiny$\pm$22} & 90{\tiny$\pm$2} & 2{\tiny$\pm$2} & 1{\tiny$\pm$1} & 55{\tiny$\pm$6} & 9{\tiny$\pm$5} & 76{\tiny$\pm$8} & 38{\tiny$\pm$18} & 83{\tiny$\pm$4} & \underline{95{\tiny$\pm$1}} & \textbf{97{\tiny$\pm$1}} \\
\midrule
\multirow{5}{*}{\texttt{ant-g}}
& 1 & 0{\tiny$\pm$0} & 0{\tiny$\pm$0} & 0{\tiny$\pm$0} & 27{\tiny$\pm$22} & 0{\tiny$\pm$0} & 0{\tiny$\pm$0} & 0{\tiny$\pm$0} & 0{\tiny$\pm$0} & 0{\tiny$\pm$1} & 0{\tiny$\pm$0} & 4{\tiny$\pm$5} & \underline{32{\tiny$\pm$12}} & \textbf{86{\tiny$\pm$3}} \\
& 2 & 0{\tiny$\pm$0} & 2{\tiny$\pm$1} & 1{\tiny$\pm$1} & 16{\tiny$\pm$17} & 0{\tiny$\pm$0} & 0{\tiny$\pm$0} & 0{\tiny$\pm$0} & 0{\tiny$\pm$0} & 4{\tiny$\pm$7} & 0{\tiny$\pm$0} & 9{\tiny$\pm$7} & \underline{63{\tiny$\pm$7}} & \textbf{92{\tiny$\pm$3}} \\
& 3 & 0{\tiny$\pm$0} & 0{\tiny$\pm$0} & 0{\tiny$\pm$0} & \underline{34{\tiny$\pm$22}} & 0{\tiny$\pm$0} & 0{\tiny$\pm$0} & 0{\tiny$\pm$0} & 0{\tiny$\pm$0} & 0{\tiny$\pm$0} & 0{\tiny$\pm$0} & 0{\tiny$\pm$1} & 1{\tiny$\pm$1} & \textbf{56{\tiny$\pm$10}} \\
& 4 & 0{\tiny$\pm$0} & 3{\tiny$\pm$0} & 0{\tiny$\pm$0} & 5{\tiny$\pm$12} & 0{\tiny$\pm$0} & 0{\tiny$\pm$0} & 0{\tiny$\pm$0} & 0{\tiny$\pm$0} & 9{\tiny$\pm$4} & 0{\tiny$\pm$0} & 14{\tiny$\pm$23} & \underline{58{\tiny$\pm$3}} & \textbf{86{\tiny$\pm$1}} \\
& 5 & 1{\tiny$\pm$1} & 12{\tiny$\pm$2} & 19{\tiny$\pm$7} & \underline{49{\tiny$\pm$22}} & 0{\tiny$\pm$1} & 0{\tiny$\pm$0} & 0{\tiny$\pm$0} & 0{\tiny$\pm$0} & 6{\tiny$\pm$10} & 13{\tiny$\pm$9} & 16{\tiny$\pm$28} & 20{\tiny$\pm$5} & \textbf{72{\tiny$\pm$3}} \\
\midrule
\multirow{5}{*}{\texttt{hum-m}}
& 1 & 1{\tiny$\pm$0} & 10{\tiny$\pm$15} & 32{\tiny$\pm$7} & 16{\tiny$\pm$9} & 1{\tiny$\pm$1} & 0{\tiny$\pm$0} & 38{\tiny$\pm$19} & 6{\tiny$\pm$2} & 25{\tiny$\pm$8} & 69{\tiny$\pm$19} & 19{\tiny$\pm$12} & \underline{78{\tiny$\pm$5}} & \textbf{93{\tiny$\pm$3}} \\
& 2 & 1{\tiny$\pm$0} & 21{\tiny$\pm$34} & 41{\tiny$\pm$9} & 18{\tiny$\pm$16} & 1{\tiny$\pm$1} & 1{\tiny$\pm$1} & 47{\tiny$\pm$35} & 40{\tiny$\pm$2} & 76{\tiny$\pm$10} & 85{\tiny$\pm$11} & \underline{94{\tiny$\pm$3}} & \underline{94{\tiny$\pm$2}} & \textbf{96{\tiny$\pm$2}} \\
& 3 & 6{\tiny$\pm$2} & 25{\tiny$\pm$31} & 25{\tiny$\pm$5} & 36{\tiny$\pm$13} & 0{\tiny$\pm$1} & 2{\tiny$\pm$1} & \underline{83{\tiny$\pm$18}} & 19{\tiny$\pm$2} & 27{\tiny$\pm$11} & 49{\tiny$\pm$49} & 74{\tiny$\pm$18} & 72{\tiny$\pm$4} & \textbf{96{\tiny$\pm$3}} \\
& 4 & 0{\tiny$\pm$0} & 5{\tiny$\pm$4} & 0{\tiny$\pm$1} & \underline{15{\tiny$\pm$16}} & 1{\tiny$\pm$1} & 1{\tiny$\pm$1} & 5{\tiny$\pm$4} & 1{\tiny$\pm$1} & 1{\tiny$\pm$2} & 1{\tiny$\pm$1} & 3{\tiny$\pm$4} & \textbf{67{\tiny$\pm$28}} & 0{\tiny$\pm$0} \\
& 5 & 2{\tiny$\pm$1} & 1{\tiny$\pm$1} & 66{\tiny$\pm$4} & 24{\tiny$\pm$20} & 1{\tiny$\pm$1} & 3{\tiny$\pm$3} & 91{\tiny$\pm$5} & 31{\tiny$\pm$7} & 63{\tiny$\pm$9} & \textbf{98{\tiny$\pm$2}} & \underline{97{\tiny$\pm$2}} & 85{\tiny$\pm$11} & \textbf{98{\tiny$\pm$1}} \\
\midrule
\multirow{5}{*}{\texttt{hum-l}}
& 1 & 0{\tiny$\pm$0} & 13{\tiny$\pm$12} & 3{\tiny$\pm$1} & 2{\tiny$\pm$1} & 0{\tiny$\pm$0} & 0{\tiny$\pm$0} & 1{\tiny$\pm$1} & 0{\tiny$\pm$0} & 0{\tiny$\pm$1} & 6{\tiny$\pm$2} & 7{\tiny$\pm$6} & \underline{57{\tiny$\pm$3}} & \textbf{79{\tiny$\pm$6}} \\
& 2 & 0{\tiny$\pm$0} & 0{\tiny$\pm$0} & 0{\tiny$\pm$0} & 0{\tiny$\pm$0} & 0{\tiny$\pm$0} & 0{\tiny$\pm$0} & 0{\tiny$\pm$0} & 0{\tiny$\pm$0} & 0{\tiny$\pm$0} & 0{\tiny$\pm$0} & 0{\tiny$\pm$0} & \underline{19{\tiny$\pm$3}} & \textbf{52{\tiny$\pm$37}} \\
& 3 & 1{\tiny$\pm$1} & 19{\tiny$\pm$7} & 7{\tiny$\pm$3} & 8{\tiny$\pm$4} & 3{\tiny$\pm$1} & 1{\tiny$\pm$1} & 2{\tiny$\pm$3} & 1{\tiny$\pm$1} & 10{\tiny$\pm$2} & 48{\tiny$\pm$10} & 11{\tiny$\pm$7} & \underline{74{\tiny$\pm$6}} & \textbf{84{\tiny$\pm$15}} \\
& 4 & 1{\tiny$\pm$0} & 5{\tiny$\pm$7} & 1{\tiny$\pm$0} & 1{\tiny$\pm$1} & 0{\tiny$\pm$0} & 0{\tiny$\pm$0} & 0{\tiny$\pm$1} & 0{\tiny$\pm$0} & 0{\tiny$\pm$0} & 1{\tiny$\pm$1} & 2{\tiny$\pm$3} & \textbf{53{\tiny$\pm$3}} & \underline{51{\tiny$\pm$37}} \\
& 5 & 0{\tiny$\pm$1} & 3{\tiny$\pm$2} & 1{\tiny$\pm$1} & 2{\tiny$\pm$2} & 0{\tiny$\pm$0} & 0{\tiny$\pm$0} & 0{\tiny$\pm$0} & 0{\tiny$\pm$0} & 1{\tiny$\pm$1} & 0{\tiny$\pm$0} & 1{\tiny$\pm$3} & \underline{47{\tiny$\pm$4}} & \textbf{77{\tiny$\pm$5}} \\
\midrule
\multirow{5}{*}{\texttt{scene}}
& 1 & 19{\tiny$\pm$6} & 0{\tiny$\pm$1} & 94{\tiny$\pm$3} & 95{\tiny$\pm$2} & \textbf{100{\tiny$\pm$0}} & 94{\tiny$\pm$4} & \textbf{100{\tiny$\pm$1}} & 87{\tiny$\pm$8} & 96{\tiny$\pm$8} & \underline{98{\tiny$\pm$3}} & \textbf{100{\tiny$\pm$0}} & \underline{98{\tiny$\pm$1}} & 96{\tiny$\pm$3} \\
& 2 & 1{\tiny$\pm$1} & 0{\tiny$\pm$0} & 12{\tiny$\pm$3} & \underline{50{\tiny$\pm$13}} & 33{\tiny$\pm$14} & 2{\tiny$\pm$2} & \underline{50{\tiny$\pm$40}} & 18{\tiny$\pm$8} & 46{\tiny$\pm$10} & 0{\tiny$\pm$0} & \textbf{76{\tiny$\pm$9}} & 12{\tiny$\pm$10} & 36{\tiny$\pm$14} \\
& 3 & 1{\tiny$\pm$1} & 0{\tiny$\pm$0} & 32{\tiny$\pm$7} & 55{\tiny$\pm$16} & \underline{94{\tiny$\pm$4}} & 4{\tiny$\pm$4} & 49{\tiny$\pm$16} & 38{\tiny$\pm$9} & 78{\tiny$\pm$14} & 54{\tiny$\pm$19} & \textbf{98{\tiny$\pm$1}} & 37{\tiny$\pm$5} & 75{\tiny$\pm$3} \\
& 4 & 2{\tiny$\pm$2} & 0{\tiny$\pm$0} & 0{\tiny$\pm$1} & 3{\tiny$\pm$3} & 4{\tiny$\pm$3} & 0{\tiny$\pm$0} & 0{\tiny$\pm$0} & \textbf{6{\tiny$\pm$1}} & 4{\tiny$\pm$4} & 0{\tiny$\pm$0} & \underline{5{\tiny$\pm$1}} & 0{\tiny$\pm$0} & 0{\tiny$\pm$0} \\
& 5 & \textbf{0{\tiny$\pm$0}} & \textbf{0{\tiny$\pm$0}} & \textbf{0{\tiny$\pm$0}} & \textbf{0{\tiny$\pm$0}} & \textbf{0{\tiny$\pm$0}} & \textbf{0{\tiny$\pm$0}} & \textbf{0{\tiny$\pm$0}} & \textbf{0{\tiny$\pm$0}} & \textbf{0{\tiny$\pm$0}} & \textbf{0{\tiny$\pm$0}} & \textbf{0{\tiny$\pm$0}} & \textbf{0{\tiny$\pm$0}} & \textbf{0{\tiny$\pm$0}} \\
\midrule
\multirow{5}{*}{\texttt{p-3x3}}
& 1 & 5{\tiny$\pm$2} & 32{\tiny$\pm$3} & 33{\tiny$\pm$6} & \textbf{97{\tiny$\pm$4}} & 52{\tiny$\pm$12} & 89{\tiny$\pm$5} & \textbf{97{\tiny$\pm$2}} & 25{\tiny$\pm$9} & 63{\tiny$\pm$19} & \underline{94{\tiny$\pm$3}} & 90{\tiny$\pm$4} & 88{\tiny$\pm$7} & \textbf{97{\tiny$\pm$5}} \\
& 2 & 1{\tiny$\pm$1} & 0{\tiny$\pm$0} & 4{\tiny$\pm$3} & 1{\tiny$\pm$1} & 0{\tiny$\pm$1} & 0{\tiny$\pm$1} & 0{\tiny$\pm$0} & 4{\tiny$\pm$2} & 2{\tiny$\pm$2} & 1{\tiny$\pm$2} & 16{\tiny$\pm$5} & \underline{74{\tiny$\pm$24}} & \textbf{96{\tiny$\pm$7}} \\
& 3 & 1{\tiny$\pm$1} & 0{\tiny$\pm$0} & 3{\tiny$\pm$2} & 3{\tiny$\pm$1} & 0{\tiny$\pm$0} & 0{\tiny$\pm$0} & 0{\tiny$\pm$0} & 1{\tiny$\pm$0} & 1{\tiny$\pm$1} & 0{\tiny$\pm$0} & 10{\tiny$\pm$3} & \underline{48{\tiny$\pm$6}} & \textbf{94{\tiny$\pm$3}} \\
& 4 & 1{\tiny$\pm$1} & 0{\tiny$\pm$0} & 2{\tiny$\pm$1} & 2{\tiny$\pm$1} & 0{\tiny$\pm$0} & 0{\tiny$\pm$0} & 0{\tiny$\pm$0} & 1{\tiny$\pm$1} & 2{\tiny$\pm$2} & 0{\tiny$\pm$0} & 16{\tiny$\pm$5} & \underline{45{\tiny$\pm$2}} & \textbf{94{\tiny$\pm$0}} \\
& 5 & 1{\tiny$\pm$0} & 0{\tiny$\pm$1} & 3{\tiny$\pm$2} & 5{\tiny$\pm$3} & 0{\tiny$\pm$0} & 0{\tiny$\pm$0} & 0{\tiny$\pm$0} & 1{\tiny$\pm$1} & 2{\tiny$\pm$2} & 0{\tiny$\pm$0} & 16{\tiny$\pm$3} & \underline{55{\tiny$\pm$35}} & \textbf{95{\tiny$\pm$3}} \\
\midrule
\multirow{5}{*}{\texttt{p-4x4}}
& 1 & 1{\tiny$\pm$1} & 9{\tiny$\pm$5} & 12{\tiny$\pm$2} & 26{\tiny$\pm$4} & 48{\tiny$\pm$5} & 24{\tiny$\pm$9} & 44{\tiny$\pm$10} & 1{\tiny$\pm$2} & 32{\tiny$\pm$9} & \underline{49{\tiny$\pm$9}} & 34{\tiny$\pm$8} & 12{\tiny$\pm$0} & \textbf{55{\tiny$\pm$6}} \\
& 2 & 0{\tiny$\pm$0} & 0{\tiny$\pm$0} & 7{\tiny$\pm$4} & 12{\tiny$\pm$4} & 14{\tiny$\pm$5} & 0{\tiny$\pm$1} & 0{\tiny$\pm$0} & 0{\tiny$\pm$1} & 5{\tiny$\pm$3} & 4{\tiny$\pm$4} & \underline{16{\tiny$\pm$5}} & 13{\tiny$\pm$5} & \textbf{20{\tiny$\pm$8}} \\
& 3 & 0{\tiny$\pm$0} & 1{\tiny$\pm$1} & 9{\tiny$\pm$3} & 15{\tiny$\pm$3} & 34{\tiny$\pm$5} & 21{\tiny$\pm$10} & 29{\tiny$\pm$12} & 1{\tiny$\pm$1} & 20{\tiny$\pm$10} & \underline{50{\tiny$\pm$14}} & 18{\tiny$\pm$5} & 32{\tiny$\pm$8} & \textbf{53{\tiny$\pm$10}} \\
& 4 & 0{\tiny$\pm$0} & 0{\tiny$\pm$1} & 5{\tiny$\pm$2} & 10{\tiny$\pm$3} & \textbf{26{\tiny$\pm$6}} & 7{\tiny$\pm$4} & 1{\tiny$\pm$1} & 0{\tiny$\pm$0} & 5{\tiny$\pm$1} & 21{\tiny$\pm$11} & 11{\tiny$\pm$3} & 12{\tiny$\pm$9} & \underline{25{\tiny$\pm$5}} \\
& 5 & 0{\tiny$\pm$0} & 0{\tiny$\pm$0} & 4{\tiny$\pm$1} & 7{\tiny$\pm$3} & \textbf{24{\tiny$\pm$11}} & 1{\tiny$\pm$1} & 0{\tiny$\pm$0} & 0{\tiny$\pm$1} & 4{\tiny$\pm$3} & 2{\tiny$\pm$2} & 7{\tiny$\pm$3} & 4{\tiny$\pm$2} & \underline{9{\tiny$\pm$4}} \\
\bottomrule
\end{tabular}%
}
\end{table*}
\appsection{OGBench Benchmark Details}
\label{appendix:benchmark}

\begin{figure}[h]
    \centering
    \includegraphics[width=0.8\columnwidth]{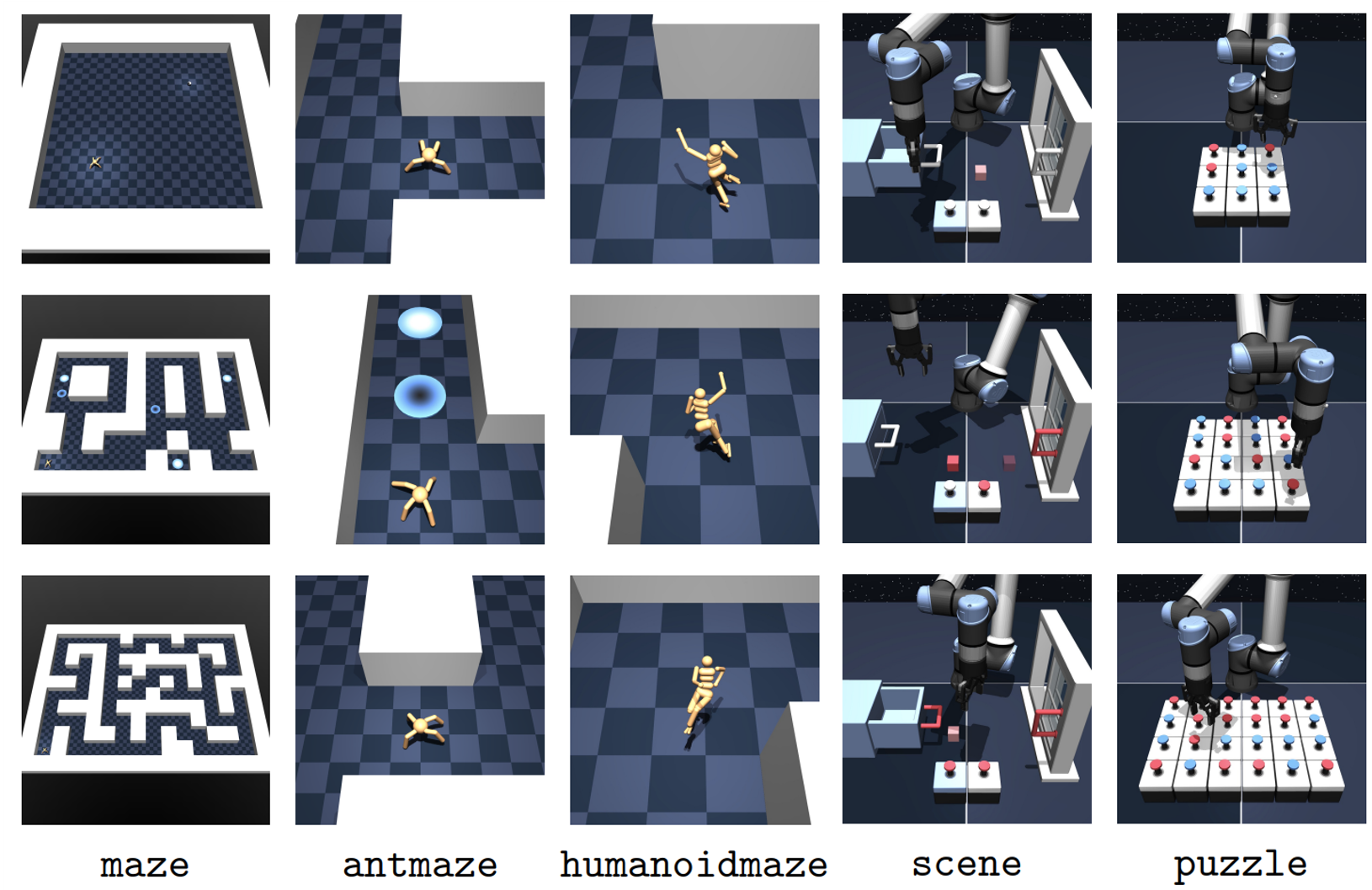}
    \caption{The $7$ OGBench environments used in our experiments.}
    \label{fig:ogbench_envs}
    \par\vspace{-0.6em}
\end{figure}

We evaluate on $7$ OGBench environments~\cite{park2025ogbench}
across three categories, with $5$ goals
(\texttt{task1}--\texttt{task5}) per environment, yielding
\rev{$35$} tasks in total. Figure~\ref{fig:ogbench_envs}
visualizes the suite.

\paragraph{Locomotion navigation}
\texttt{ant-large} and \texttt{ant-giant} use an $8$-DoF quadruped,
while \texttt{hum-medium} and \texttt{hum-large} use a $21$-DoF
whole-body humanoid in the same maze layouts. The \texttt{giant}
variants require longer horizons, while the humanoid tasks have
higher-dimensional actions and richer multimodal behavior, where
generative actors are particularly beneficial~\cite{park2025fql}.

\par\vspace{-0.6em}
\paragraph{Multi-step manipulation}
\texttt{scene} requires sequences of pick-place and tool-use
sub-actions under sparse rewards, demanding temporally extended
planning.

\par\vspace{-0.6em}
\paragraph{Combinatorial-stitching manipulation}
\texttt{puzzle-3x3} and \texttt{puzzle- 4x4} require solving
sliding-tile puzzles on $3{\times}3$ and $4{\times}4$ grids,
combining continuous control with discrete combinatorial structure
and many behavioral modes. \texttt{puzzle-4x4} is among the hardest
tasks in the suite.

\appsection{Experimental Details}
\label{appendix:hyperparameters}

Table~\ref{tab:hyperparams} lists the architecture and training
hyperparameters of LAC. Unless otherwise noted, the same values are
used across all $7$ OGBench environments.

\paragraph{\rev{Bootstrap sampling}}
\rev{$n$-step returns $G_t^{(n)} = \sum_{k=0}^{n-1} \gamma^{k} r_{t+k}$
are computed on the fly in a vectorized pass at sampling time.
Windows never cross trajectory ends, and the bootstrap term is
zeroed whenever a terminal state falls inside the window.}

\paragraph{Target networks}
Both the critic target $Q_{\theta^-}$ and the actor target
$\pi_{\phi^-}$ are maintained by Polyak averaging at every gradient
step. The actor target is used only to compute the next-state
action that enters the $n$-step bootstrap target, following the
standard practice of TD3-style methods~\cite{fujimoto2021minimalist}.

\paragraph{Evaluation protocol}
We follow the OGBench
protocol~\cite{park2025ogbench,park2025fql}: success rates are
averaged across the last three evaluation epochs and over $4$ random seeds, reducing single-epoch
noise.

\paragraph{Hardware (training)}
All training runs are performed on a single NVIDIA RTX PRO 6000
GPU. A single training run of LAC takes approximately $1$ hour;
the full set of \rev{$35$} tasks under $4$ seeds amounts to roughly \rev{$6$}
GPU-days.

\paragraph{Hardware (inference latency)}
For controlled comparability with prior reports on consumer-grade
hardware, the per-action latency measurements in
Table~\ref{tab:latency} are obtained on a separate NVIDIA RTX 3090
GPU. Each measurement uses batch size $1$, $100$ warm-up iterations
to amortize JIT compilation, and is averaged over $10{,}000$
measurement iterations. All baselines are measured under an
identical protocol on the same hardware.

\par\vspace{-0.6em}
\appsection{Full OGBench Results}
\label{appendix:ogbench_full}
Table~\ref{tab:ogbench_full} reports per-task success rates for 
all \rev{$35$} OGBench tasks, covering both locomotion 
(\texttt{antmaze}, \texttt{humanoidmaze}) and manipulation 
(\texttt{scene}, \texttt{puzzle}) domains. Each 
cell reports the mean and standard deviation over $4$ random 
seeds, with $50$ evaluation episodes per seed at the last three evaluation epochs. Baseline numbers other than TD3+BC are taken 
directly from the FQL paper~\cite{park2025fql} for fair 
comparison; TD3+BC is re-run under our codebase to ensure direct 
comparability with our deterministic-actor configurations 
(\textsc{LAC-S}, \textsc{LAC-L}).

\par\vspace{-0.6em}
\section*{Acknowledgment}
This work was partly supported by the Institute of Information \& Communications Technology Planning \& Evaluation (IITP) grant (No. RS-2026-25519475), the National Research Foundation of Korea (NRF) grants (RS-2023-00278812, RS-2025-02214082, RS-2026-25522801) funded by the Korean government (MSIT), and Korea Institute of Police Technology (KIPoT) funded by the Korean National Police Agency \& Korea Customs Service (RS-2026-25536784). The authors would like to thank Chanin Eom for his helpful discussions and comments during this research.


\section*{Generative AI Usage Disclosure}
We disclose how generative AI tools were used in the preparation 
of this work.
\begin{itemize}
    \item \textbf{Manuscript preparation:} Large language model 
    assistants were used to polish English phrasing and improve 
    readability.
    \item \textbf{Implementation:} Large language model assistants 
    were used to support coding and debugging during development of 
    the LAC codebase.
    \item \textbf{Experiments and analysis:} All training runs, 
    metric computations, and interpretations of the empirical 
    findings were conducted solely by the authors.
\end{itemize}


\end{appendices}
\bibliographystyle{ACM-Reference-Format}
\bibliography{LAC_ref}

\end{document}